\documentclass{article}

\PassOptionsToPackage{numbers,sort&compress}{natbib}

\usepackage[preprint]{neurips_2026}
\usepackage{amsmath}
\usepackage{graphicx}

\usepackage[utf8]{inputenc} 
\usepackage[T1]{fontenc}    
\usepackage{hyperref}       
\usepackage{url}            
\usepackage{booktabs}       
\usepackage{amsfonts}       
\usepackage{nicefrac}       
\usepackage{microtype}      
\usepackage{xcolor}         

\workshoptitle{Responsible Communication of Machine Learning Research in Biomedicine}
\title{From Confidence Scores to Decision Consequences: A Decision-Theoretic Approach to Communicating Uncertainty for Atrial-Fibrillation Detection%
\thanks{Accepted at the NeurIPS 2026 Workshop on Responsible Communication of Machine Learning Research in Biomedicine.}}

\author{%
  Ciaran Bench\\
  Department of Data Science and AI\\
  National Physical Laboratory\\
  Teddington, UK \\
  \texttt{ciaran.bench@npl.co.uk} \\
}

\begin{document}

\maketitle

\begin{abstract}
While deep learning models have significant potential to improve treatment outcomes, the risk of poor performance on previously unseen data hinders their deployment in practical settings. Uncertainty quantification (UQ) in deep learning may provide an indication of the trustworthiness of model predictions. However, the practical utility of uncertainties for detecting outputs that may negatively impact clinical workflows is often unclear given common UQ evaluation frameworks do not convey the consequences of acting on a model prediction. We investigate decision-theoretic uncertainty as a mechanism for communicating the consequences of using a model's atrial fibrillation classification prediction. Rather than treating confidence scores as an indicator of accuracy, we express Venn-Abers recalibrated confidence scores as expected costs under a toy decision-cost model inspired by screening pathways for wearable devices. We then evaluate when this decision-theoretic representation supports prediction filtering, how its utility changes with asymmetric error costs, and whether its relationship with realised decision-cost remains reliable under synthetic noise-induced domain shift. Cost-aware filtering generally achieved lower realised decision cost at higher retain fractions when the costs of false-positive and false-negative decisions were highly asymmetric; however, its benefit varies with shift severity. This work supports a distinction between communicating predictive doubt from decision consequences.  
  \end{abstract}

\section{Introduction}
While deep learning has considerable potential to improve healthcare outcomes, the risk of poor performance on previously unseen data can hinder the deployment of models in more practical settings. Uncertainty quantification (UQ) in deep learning aims to provide an indication of the doubt in a given prediction, which can be used to reject predictions likely to be incorrect (among other uses) \cite{abdar2021review}. However, estimating these quantities is challenging.

Among several concepts usually considered in the evaluation of the quality/utility of predicted uncertainties, calibration is one important aspect that, for classification, assesses whether the magnitude of the predicted uncertainty correlates with the empirical frequency of outcomes \cite{guo2017calibration}. For classification tasks, one expression of uncertainty is the post-softmax pseudoprobability for the predicted class, referred to as a confidence score. Calibration may be evaluated by comparing binned confidence scores against the corresponding accuracy rate for each bin (e.g. through the Expected Calibration Error (ECE) \cite{guo2017calibration}). However, this does not encode the consequences of using predictions in a given decision-making process \cite{van2025evaluation}. For example, in a given scenario it may be that false positives incur higher decision costs (negative consequences) than false negatives. Consequently, the practical insights about the utility of estimates in realistic deployment scenarios through generic notions of confidence calibration are limited. 

Recent works have emphasised the need to use expressions of uncertainty that provide more actionable and interpretable insights to users \cite{harrigan2022considerations,frans2023visualizing,banerji2023clinical,flores2026aligning,cortesgomez2025utility}. This includes emerging interest in decision-theoretic UQ \cite{smith2024rethinking,canal2024decision}, which aims to express uncertainty through the expected decision cost of using predictions. Here, uncertainties may be used with a decision policy to select the action associated with a given prediction, where the cost of the action is defined by an \textit{expected} cost function. This framework may provide a more decision-relevant form of the expected consequences of using a prediction in a decision-making process. 

The corresponding evaluation of the quality of decision-theoretic uncertainties aims to show whether using model predictions with lower estimated uncertainty corresponds with lower \textit{realised} decision cost (i.e. the actual cost incurred from using predictions) \cite{smith2024rethinking}. In the context of deep learning, this framework has been used for large language models as a form of subjective UQ, which is appealing given the lack of ground truths for generated outputs \cite{wang2024subjective}. Applications in medical deep learning that involve decision-theoretic expressions of uncertainty are comparatively limited, where \cite{kwon2026conformal} investigated the use of expected decision cost to inform the choice of decision threshold for a classification model. However, decision-theoretic principles have been considered in formulating strategies for using model predictions \cite{vickers2016net}.  
\subsection{Aims}
Here, we apply the decision-theoretic framework to a distinct use case where uncertainties are used to \textit{filter out} predictions likely to incur high decision cost rather than only as a means of selecting a single operating threshold. Here, uncertainty is used as a direct indicator of the expected utility of model predictions. In other works, cost models have been employed to facilitate accept/reject filtering for classification model predictions \cite{franc2021optimalstrategiesrejectoption}. However, the utility of this approach under realistic deployment settings (e.g. under test time domain shift), and in biomedical applications of deep learning (which may have unique characteristics to the cost functions) remain relatively uncharacterised. 

We address this gap through a  case study in atrial fibrillation classification. \textbf{We evaluate whether Venn-Abers recalibrated confidence scores re-expressed as \textit{expected} decision costs can help identify predictions with high \textit{realised} cost and therefore support selective prediction under dataset shift.} We consider three variants of an economic decision cost model inspired by clinical workflows, each with progressively higher asymmetry in the cost associated with positive and negative prediction. We do not establish that decision-theoretic uncertainty improves clinician comprehension. Instead, we evaluate whether the quantity being communicated remains empirically connected to the decision consequence it represents, including under domain shift.

\subsection{Venn-Abers confidence recalibration under domain shift}

Venn-Abers recalibration is an appealing conformal-style calibration framework that we consider here as it has precedent in the medical machine learning literature \cite{pereira2020targeting,schmidt2026machine,bench2026systematic}, and because it is sensitive to domain shift, providing a convenient means to study the utility of decision-theoretic uncertainty under realistic test-time conditions \cite{vovk2012venn}. Here, a hold out calibration-set (separate from the train/validation/test splits) is used to predict probability intervals at test time. If the calibration set is exchangeable with the test set, the resulting recalibrated test set confidence scores (i.e. those acquired by collapsing the intervals into a single value) will have validity guarantees (i.e. the recalibrated confidence reflects the empirical frequency that the ground truth and the predicted class are the same). 

Following the procedure described in \cite{vovk2012venn}, given a pretrained classifier $s$, a hold out calibration set of inputs $X^c = x^c_1,x^c_2,... x^c_i,...x^c_N$, and ground truths $Y^c = y^c_1,y^c_2,... y^c_i, ... y^c_N$, and a hold out test set $X^t = x^t_1,x^t_2,... x^t_j,...x^t_M$, and $Y^t = y^t_1,y^t_2,... y^t_j,...y^t_M$, for each test input $x^t_j$ one additional dataset is formed by adding the example $(s(x^t_j),0)$ to the set of predictions  ${s(X^c),Y^c}$, while another  dataset is formed by adding the example $(s(x^t_j),1)$ to the set of predictions  ${s(X^c),Y^c}$. A separate isotonic regression model $g^0_j$ and $g^1_j$ is trained on each of these datasets. The prediction interval for $x^t_j$ is given by $[\min (g^0_j(x_j),g^1_j(x_j)),\max (g^0_j(x_j),g^1_j(x_j))]$. Here, we use the default parameterisation of the \verb |sklearn.isotonic.IsotonicRegression| regression function. By minimising the worst-case logarithmic loss over all true probabilities consistent with the Venn prediction, we can define a single probability as $\hat{p} = \frac{p_1}{1 - p_0 + p_1}$, where $p_0 = \min\!\left(g^0_j(x_j),g^1_j(x_j)\right)$ and $p_1 = \max\!\left(g^0_j(x_j),g^1_j(x_j)\right)$. We consider our expression of uncertainty as this recalibrated confidence.

The utility of the recalibrated probabilities in deployment settings depends on whether they encode information related to decision consequences, and whether these indicators remain reliable when processing previously unseen data. This second condition motivates the need to evaluate uncertainty reliability/calibration under dataset shift. Some works have shown that uncertainties predicted from a wide range of UQ techniques do not generalise to out of distribution data \cite{ovadia2019can,hekler2025beyond,slavutsky2026quantifying,NEURIPS2019_8fb21ee7}. However, these analyses do not consider realistic cost models, and instead rely on more generic indicators of calibration. These do not consider the realised consequences of using predictions, and instead capture concepts like coverage, sharpness, or calibration. Therefore, the practical impact of distribution shift with respect to how uncertainties and model outputs are used in practice is unclear. This limits the practical utility of these analyses and the evaluation approach. Here we impose domain shift through the addition of random noise to input PPG time series.

\section{Methods}
\subsection{Atrial Fibrillation classification}
We consider a case study of the classification of atrial fibrillation (AF, periods of rapid uncoordinated heartbeat) from wearable PPG time series \cite{wasserlauf2019smartwatch}. PPG is an optical sensing modality that measures changes in blood flow at the vascular periphery (e.g. at the fingertip, or earlobe). Here, near-infrared optical light is shone through the tissue at different wavelengths, where variations in the amount transmitted/backscattered through the tissue due to changes in the tissue optical properties spurred by pulsatile flow are measured and encoded in 1D time series \cite{charlton2022wearable}. Information about blood flow is related to physiological parameters of interest, such as heart rate, which makes this mode of measurement appealing for detecting AF. The recent emergence of wearable PPG sensing technologies has created the opportunity for continuous monitoring. The large quantities of data generated and the need for ongoing evaluation make deep learning models a compelling framework for this application \cite{torres2020multi}.

\subsection{Realised decision cost}
\label{sec:realisedcost}
Here, the utility of the uncertainties is evaluated based on the \textit{realised} decision cost. The realised decision cost provides the average per-example economic cost of following the screening pathway associated with each prediction. The screening pathway for a given prediction is determined by the probability threshold $\tau$ ($\hat{p}\geq\tau$ corresponds to positive AF prediction), filtering rule (how sufficiently high uncertainty predictions are determined), and the reject pathway (how filtered predictions are incorporated into the cost evaluation).

While there are several notions of decision cost that could be considered in our evaluation, (e.g. lives saved, quality of life improvement, or comfort/efficiency of care) we consider monetary cost given the prevalence of this choice in the AF screening literature \cite{welton2017methods,hill2020identification,chen2022cost}, and because it is one of the main motivations for the use of AI models in health. Unlike other AF screening studies such as \cite{chen2022cost} that consider Markov or microsimulation models to capture stroke, bleeding, mortality, and quality-adjusted life-years over time, our formulation considers a ``static" expected cost over one year. A more careful treatment may consider the costs of missed diagnosis over short-term monitoring windows involving continuously repeated measurements; though limitations in the length of the streams here prevent an effective analysis. Therefore, our cost formulation should be interpreted as a clinically motivated reduced-form, static proxy for decision cost inspired by the diagnostic workflow described in \cite{chen2022cost}. This is a toy model that enables the comparison of prediction filtering strategies under varying degrees of asymmetry in positive and negative costs, rather than a comprehensive health-economic model, or one that is capable of factoring in all relevant aspects of a longitudinal/continuous monitoring scenario. The diagnostic workflow presented in \cite{chen2022cost} is appealing given it aligns with our use case of using wearable devices. Additionally, some of the parameters used in the model are acquired from external sources, and may not apply to the datasets implemented here (ISPOR good-practice guidance \cite{siebert2012state} and CHEERS \cite{husereau2022consolidated} reporting guidance both support transparent use of literature-based model parameters).

Including a fixed screening cost \(C_{\mathrm{screen}}\) incurred for each
screened individual, the toy realised cost model is defined as,

\begin{equation}
\bar C
= C_{\text{screen}}
+ \pi\left[\mathrm{Se}\,C_{\text{TP}} + (1-\mathrm{Se})\,C_{\text{FN}}\right]
+ (1-\pi)\left[(1-\mathrm{Sp})\,C_{\text{FP}} + \mathrm{Sp}\,C_{\text{TN}}\right],
\label{eq:fullrealcost}
\end{equation}

where $\pi$ denotes the prevalence of AF in the screened population, while $\mathrm{Se}$ and $\mathrm{Sp}$ denote the sensitivity and specificity of the classifier, respectively. The terms $C_{\mathrm{TP}}$, $C_{\mathrm{FN}}$, $C_{\mathrm{FP}}$, and $C_{\mathrm{TN}}$ denote the costs associated with true-positive, false-negative, false-positive, and true-negative outcomes, respectively, and $C_{\mathrm{screen}}$ is the fixed cost incurred for each screened individual. Further details of the cost components are provided in Table \ref{tab:basic_cost_parameters} in Appendix \ref{app:costterms}. The derivation for this model is given in Appendix \ref{app:derivrealcost}. Given some modelling assumptions described in Appendix \ref{app:costterms} (i.e. $C_{TP}=0$, and $C_{TN}=0$), this reduces to,
\begin{equation}
\bar C
= C_{\text{screen}}
+ \pi\left[ (1-\mathrm{Se})\,C_{\text{FN}}\right]
+ (1-\pi)\left[(1-\mathrm{Sp})\,C_{\text{FP}} \right].
\label{eq:realcost}
\end{equation}

We consider three variants of the cost model, each parameterised by different values for the probability of stroke occurring in a year, given a missed AF screening. This affects the costs of false negatives. We consider $p_{\text{stroke|missed AF}} = 0.025$ where $p_{\text{stroke|missed AF}}$ corresponds to CHA$_2$DS$_2$-VASc (a clinical risk stratification score used to estimate a patient’s annual risk of stroke with AF) of 2 at 2.5\%/year which represents a lower bound for older populations \cite{chen2022cost}\cite{johnson2016short}. We also consider  $p_{\text{stroke|missed AF}} = 0.037$ which corresponds to CHA$_2$DS$_2$-VASc of 3 at 3.7\%/year, and $p_{\text{stroke|missed AF}} = 0.055$ which corresponds to CHA$_2$DS$_2$-VASc of 4 at 5.5\%/year. This increases the cost of negative predictions relative to positive predictions. 

\subsubsection{Reject pathway}
\label{sec:rej}
Here, realised cost is computed only for those examples retained after confidence-filtering (retain-only). This retain-only pathway is relevant to the hypothetical scenario where a confidence score is used to determine the trustworthiness of a prediction, where clinical actions are only made on confident predictions. E.g., a low-confidence prediction may be rejected (i.e. not used to make a clinical action) and trigger the need to take a repeat measurement (at negligible cost) from the wearable device.

A comprehensive economic cost model would have to consider additional factors such as the likelihood that low confidence is resolved with repeat measurements (and not due to more systematic/consistent sources of error like device malfunction/user error), and that the resulting delay incurred does not significantly influence the effectiveness of follow-up treatment pathways. Modelling these outcomes requires additional empirical and clinical information that is not found in the dataset considered and is outside the scope of the present work. Therefore, the retain-only setting should be interpreted as reflecting a simplified/toy deployment scenario to help illustrate the potential strengths and additional insights a decision-centric approach to reliability analysis can provide in principle rather than a recommended clinical workflow or a fully comprehensive cost model.

\subsection{Filtering rules}
We consider two prediction filtering rules: a \textbf{na\"{i}ve} rule, and a decision-theoretic \textbf{cost-aware} rule. Both strategies enable filtering without prior knowledge of the ground truth, though the latter makes use of decision-theoretic expressions of uncertainty.

\subsubsection{ Na\"{i}ve filtering}
For the \textbf{na\"{i}ve} rule, predictions with the lowest $X\%$ confidence scores are rejected (i.e. those closest to 50\%) for several values of $X$. Here, if the recalibrated confidence score for the positive class $\hat{p}\geq\tau$, where $\tau=0.5$, then this triggers the positive AF screening pathway; otherwise, the negative screening pathway is triggered. Rejected predictions are either excluded from the cost calculation, or assigned the negative pathway depending on the reject pathway.

\subsubsection{Decision-theoretic uncertainty and cost-aware filtering}
In the \textbf{cost-aware} rule, we reject predictions whose \textit{expected decision cost} is among the top $X\%$ in the entire test dataset \cite{smith2024rethinking}. For the cost-aware rule, we define an \textit{expected} cost function inspired by Equation \ref{eq:realcost} for a \textit{single} model prediction in the context of AF screening for the cost-aware rule. This allows recalibrated confidences to be expressed as decision-theoretic indicators of trustworthiness.

Let $\hat{p} \in [0,1]$ denote the recalibrated confidence for an AF prediction, and let $\hat{y} = \mathbb{I}(\hat{p} \ge \tau)$ be the binary classification obtained by thresholding the probability with a threshold value of $\tau \in [0,1]$, where $\mathbb{I}(\cdot)$ is the indicator function. Let $a \in \{0,1\}$ denote the action taken after observing the model output (parameterised by $\tau$), where $a=1$ corresponds to triggering the AF-positive diagnostic pathway and $a=0$ to treating the individual as AF-negative.
Given a Venn-Abers calibrated probability $\hat{p}=P(y=1\mid x)$, the expected cost of the chosen action (screening pathway) is obtained by marginalising the event-specific costs over the possible true labels. The expected cost of assigning the individual to the positive class ($\hat{p}\geq\tau$) is then,
\begin{equation} 
\mathbb{E}[C \mid \hat{y}=1, \hat{p}]
=
\hat{p}\,C_{\mathrm{TP}} + (1-\hat{p})\,C_{\mathrm{FP}} + C_{\mathrm{screen}}.
\end{equation}
The expected cost of assigning the individual to the negative class ($\hat{p}<\tau$) is,
\begin{equation}
\mathbb{E}[C \mid \hat{y}=0, \hat{p}]
=
\hat{p}\,C_{\mathrm{FN}} + (1-\hat{p})\,C_{\mathrm{TN}}
+ C_{\mathrm{screen}}.
\end{equation}

For the cost-aware rule, we consider decision-optimal values for $\tau$, which correspond to the probability at which positive and negative predictions incur the same expected cost. Therefore, a different value of $\tau$ is used for each variant of the cost model ($\tau = \{0.381, 0.293, 0.218\}$ for $p_{\text{stroke}} = \{0.025, 0.037, 0.055\}$ respectively). We also provide results for using $\tau=0.5$ for the cost-aware rule in Appendix \ref{sec:tau5retain} (aligned with that used for the na\"{i}ve rule) to isolate the effect of the filtering rule more directly, and find that while the differences in cost are lower in magnitude, the qualitative insights of the study remain. 

\subsection{Model training and evaluation}
\label{sec:training}
Our aim is to evaluate how the common failure mode of dataset shift/violation of calibration-test exchangeability affects decision cost. Here, we impose dataset shift through the augmentation of the input time series with random Gaussian noise (parameterised with standard deviation of $\sigma= \{0.1, 0.3, 0.6\}$ as a means to provide both more subtle and severe corruption to the signals, which are normalised to a range of [0,1] prior to adding noise, where they are then renormalised using the min/max of the unaugmented data). We report the results for $\sigma=0.3$ in the main text, and provide other results in the Appendix (Fig. \ref{fig:cleannoisyex} shows some example time series).

While this scenario is synthetic, it provides a controllable means of introducing significant dataset shift that is convenient for this analysis, and models variations in signal quality that may emerge in realistic deployment conditions (e.g. acquisition with low quality equipment). Therefore, the insights drawn here may still generalise to other scenarios.

We train and evaluate the AF classification model following the procedure in \cite{bench2025uncertainty}. Here, a 1D variant of AlexNet is trained on a custom split of the Deepbeat dataset, using a custom loss that models distributions for each logit \cite{kendall2017uncertainties}. Here we use early stopping with a patience of 25 epochs. At evaluation, a single sample of the softmaxed logits are acquired, and the resulting confidence scores are then recalibrated using the Venn-Abers technique. We then apply the filtering rules using a range of retain fractions from 0.1 to 1 in steps of 0.05. We compute cost-coverage curves for the predicted confidences, which plot mean realised cost compared to the fraction of retained examples.

\section{Results and discussion}

\subsection{Venn-Abers ECE under domain shift}
\begin{figure}[t]
    \centering
\includegraphics[width=0.49\linewidth]{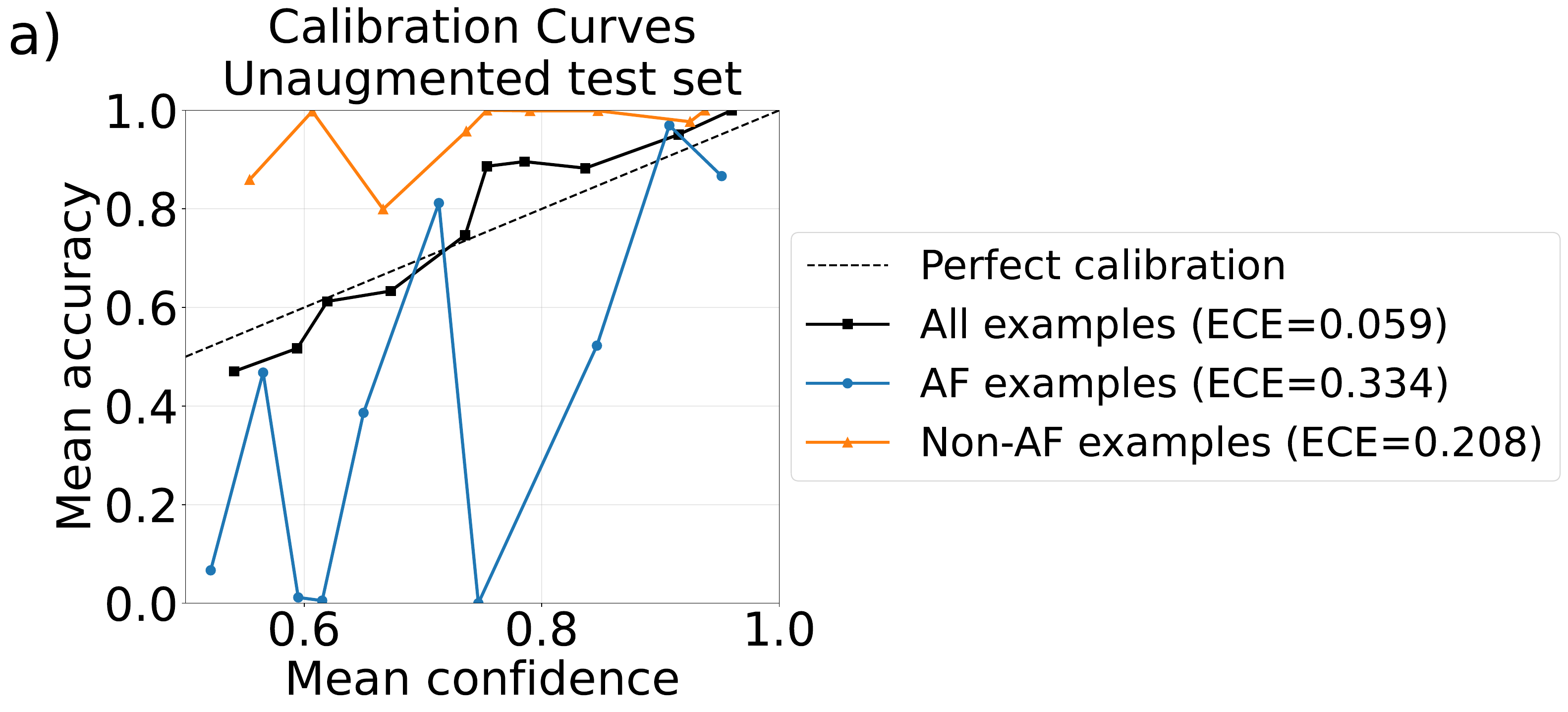}
\includegraphics[width=0.49\linewidth]{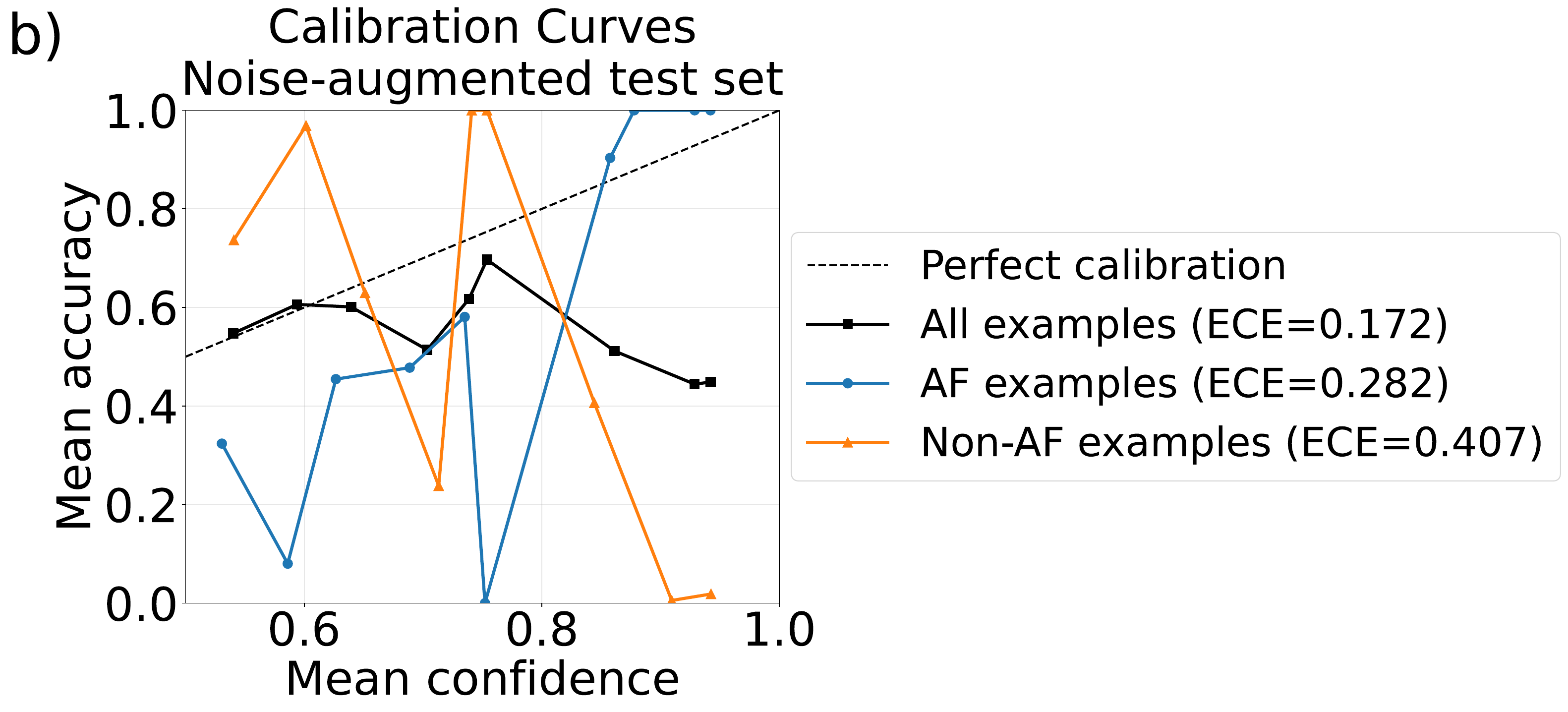}
    \caption{Calibration curves for the recalibrated test predictions for the original and noise-augmented test set (here $\sigma=0.3$ is considered). Expected Calibration Error (ECE) values are shown for full test set and class-wise splits.}
    \label{fig:calibsurves}
\end{figure}

\begin{table} \centering \scriptsize \caption{Composition of prediction outcomes for the overall test set and rejected subset at a 50\% retain fraction. N denotes the na\"{i}ve rule and C denotes the cost-aware rule. Percentages are calculated relative to the total number of examples in the corresponding subset. We use decision-optimal values of $\tau$ for the cost-aware rule and $\tau=0.5$ for the na\"{i}ve rule. The cost-aware decision thresholds are $\tau=\{0.381$, $0.293$, $0.218$\} for $p_{\mathrm{stroke}}=\{0.025$, $0.037$, $0.055\}$, respectively.} \label{tab:error-compositionnoise03} \begin{tabular}{lllrrrrr} \toprule $p_{\mathrm{stroke}}$ & Rule/data & Subset & Total & TN, n (\%) & FN, n (\%) & TP, n (\%) & FP, n (\%) \\ \midrule All & Augmented & Overall & 15377 & 5834 (37.94\%) & 2867 (18.64\%) & 2930 (19.05\%) & 3746 (24.36\%) \\ All & Original & Overall & 15377 & 9171 (59.64\%) & 3464 (22.53\%) & 2333 (15.17\%) & 409 (2.66\%) \\ \midrule All & N, augmented & Rejected & 7689 & 3295 (42.85\%) & 1860 (24.19\%) & 1114 (14.49\%) & 1420 (18.47\%) \\ All & N, original & Rejected & 7689 & 3408 (44.32\%) & 2816 (36.62\%) & 1097 (14.27\%) & 368 (4.79\%) \\ \midrule 0.025 & C, augmented & Rejected & 7689 & 2743 (35.67\%) & 1269 (16.50\%) & 1446 (18.81\%) & 2231 (29.02\%) \\ 0.025 & C, original & Rejected & 7689 & 2335 (30.37\%) & 975 (12.68\%) & 2325 (30.24\%) & 2054 (26.71\%) \\ \midrule 0.037 & C, augmented & Rejected & 7689 & 3109 (40.43\%) & 1394 (18.13\%) & 1136 (14.77\%) & 2050 (26.66\%) \\ 0.037 & C, original & Rejected & 7689 & 3177 (41.32\%) & 694 (9.03\%) & 1772 (23.05\%) & 2046 (26.61\%) \\ \midrule 0.055 & C, augmented & Rejected & 7689 & 208 (2.71\%) & 15 (0.20\%) & 2449 (31.85\%) & 5017 (65.25\%) \\ 0.055 & C, original & Rejected & 7689 & 731 (9.51\%) & 98 (1.27\%) & 1941 (25.24\%) & 4919 (63.97\%) \\ \bottomrule \end{tabular} \end{table}

\begin{table}[t] \centering \small \caption{Area under the cost--coverage curve (AUCC) for the na\"{i}ve and cost-aware filtering rules on the original and noise-augmented test sets for $\sigma=0.3$. Lower AUCC indicates lower realised cost across the evaluated retain fractions. The na\"{i}ve rule uses $\tau=0.5$, whereas the cost-aware rule uses the decision-optimal thresholds $\tau=\{0.381,0.293,0.218\}$ for $p_{\mathrm{stroke}}=\{0.025,0.037,0.055\}$, respectively.} \label{tab:auccnoise03} \begin{tabular}{llrr} \toprule $p_{\mathrm{stroke}}$ & Test set & Na\"{i}ve AUCC (\$) & Cost-aware AUCC (\$) \\ \midrule 0.025 & Original & 458.6814 & 442.6533 \\ 0.025 & Augmented & 669.0770 & 672.2776 \\ \midrule 0.037 & Original & 512.4814 & 471.1761 \\ 0.037 & Augmented & 719.9655 & 692.7153 \\ \midrule 0.055 & Original & 593.1815 & 514.7212 \\ 0.055 & Augmented & 796.2983 & 693.5372 \\ \bottomrule \end{tabular} \end{table}

The ECE of the unaugmented test set before Venn-Abers calibration was 0.087, while recalibration reduced this to 0.059, which demonstrates that recalibration did improve confidence calibration as assessed by the ECE. Under the noise-induced domain shift, however, the reliability of the recalibrated probabilities deteriorated, as shown in Fig. \ref{fig:calibsurves}. This result is consistent with the dependence of Venn-Abers recalibration on the exchangeability of the calibration and test set, and with expected degradation of confidence calibration for OOD data. However, ECE does not indicate how calibration degradation affects the consequences of decisions made downstream of the model's prediction. The same ECE may have substantially different consequences depending on the relative costs of false-positive and false-negative decisions. We therefore complement this more standard calibration analysis with a decision-theoretic analysis.

\begin{figure}[t]
    \centering
    \includegraphics[width=0.45\linewidth]{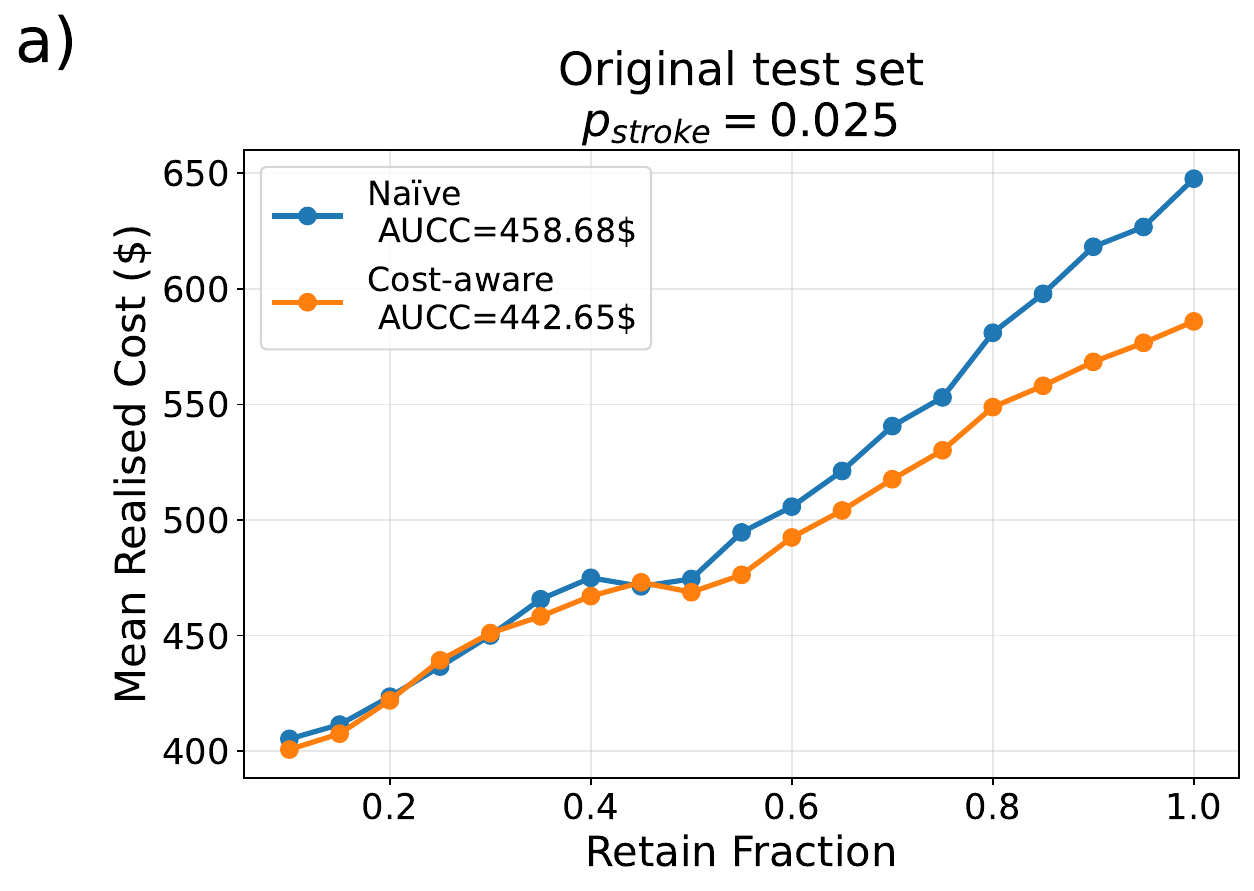}\includegraphics[width=0.45\linewidth]{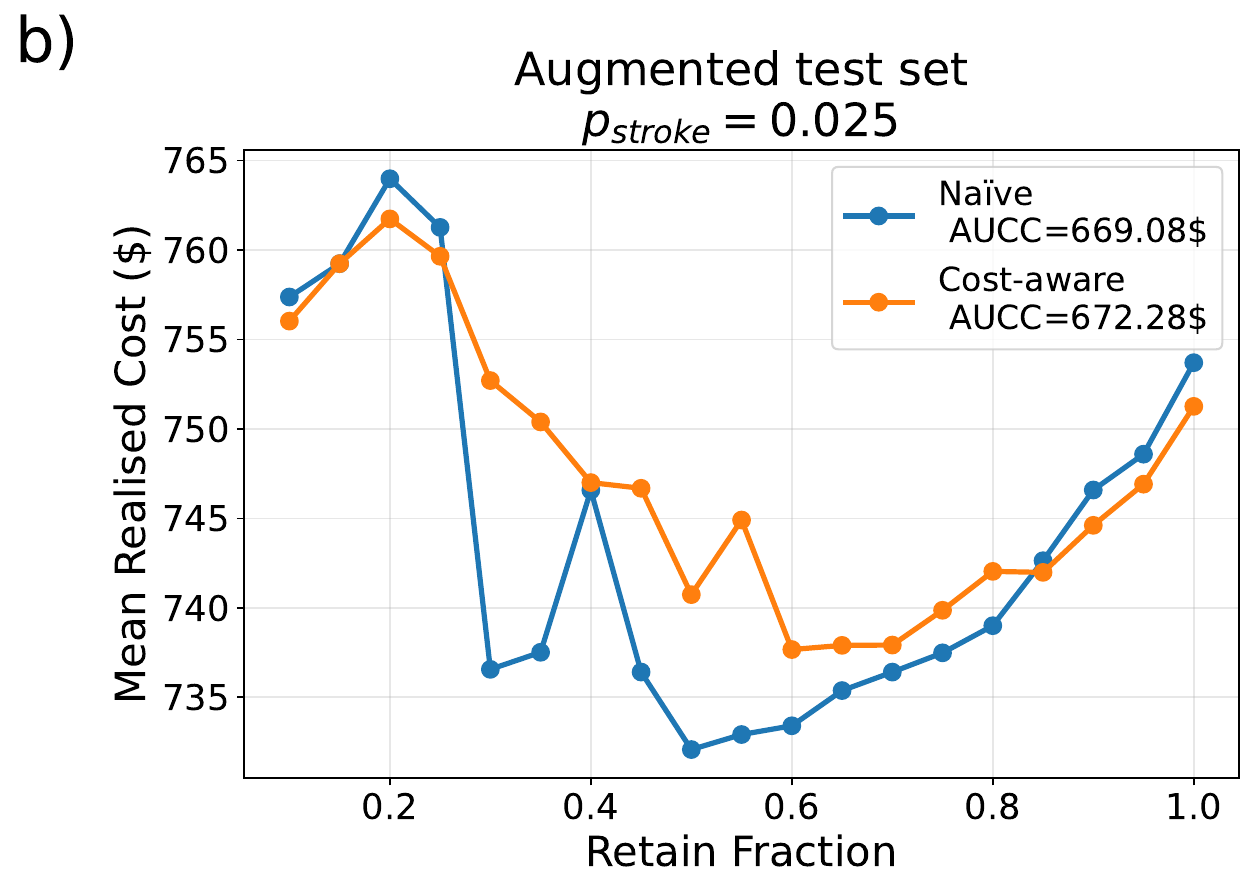}
    \includegraphics[width=0.45\linewidth]{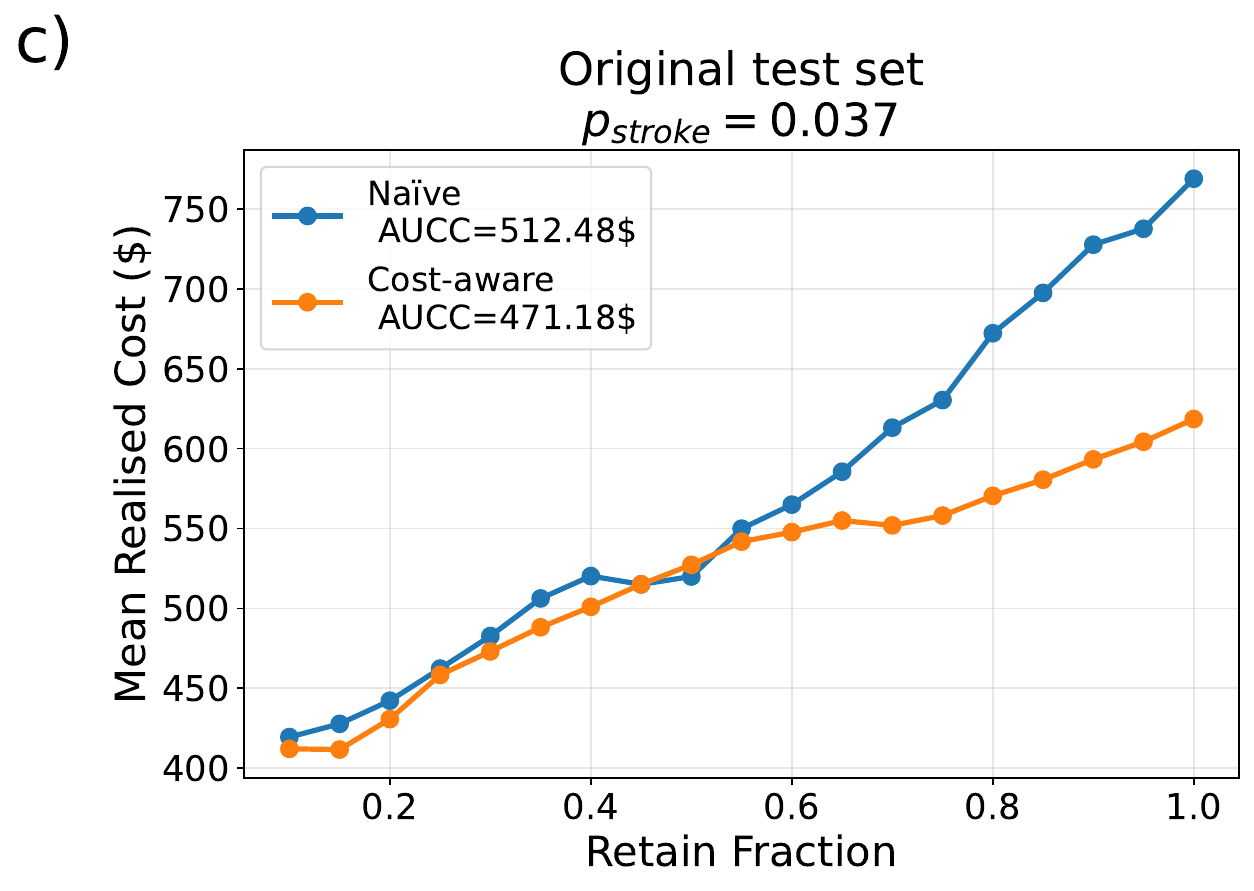}
    \includegraphics[width=0.45\linewidth]{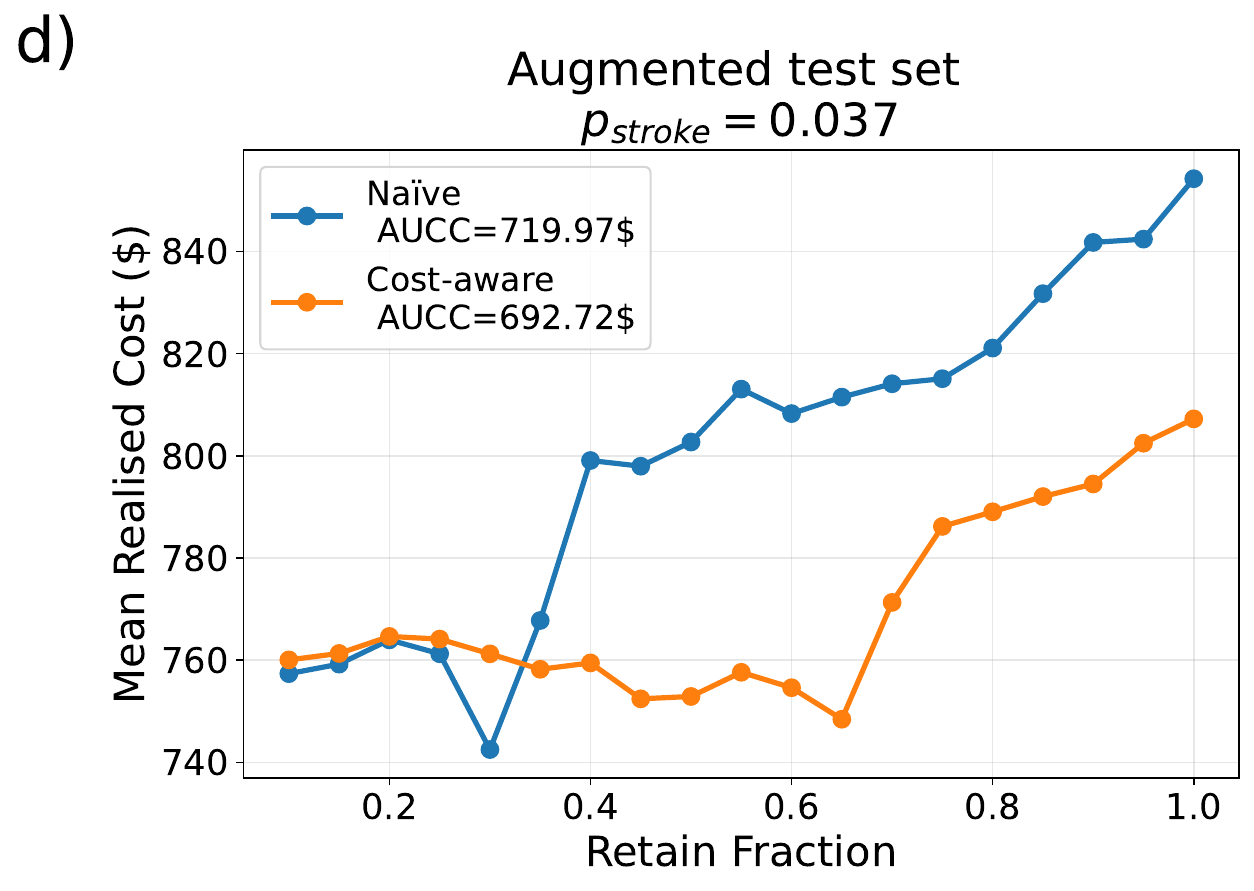}
    \includegraphics[width=0.45\linewidth]{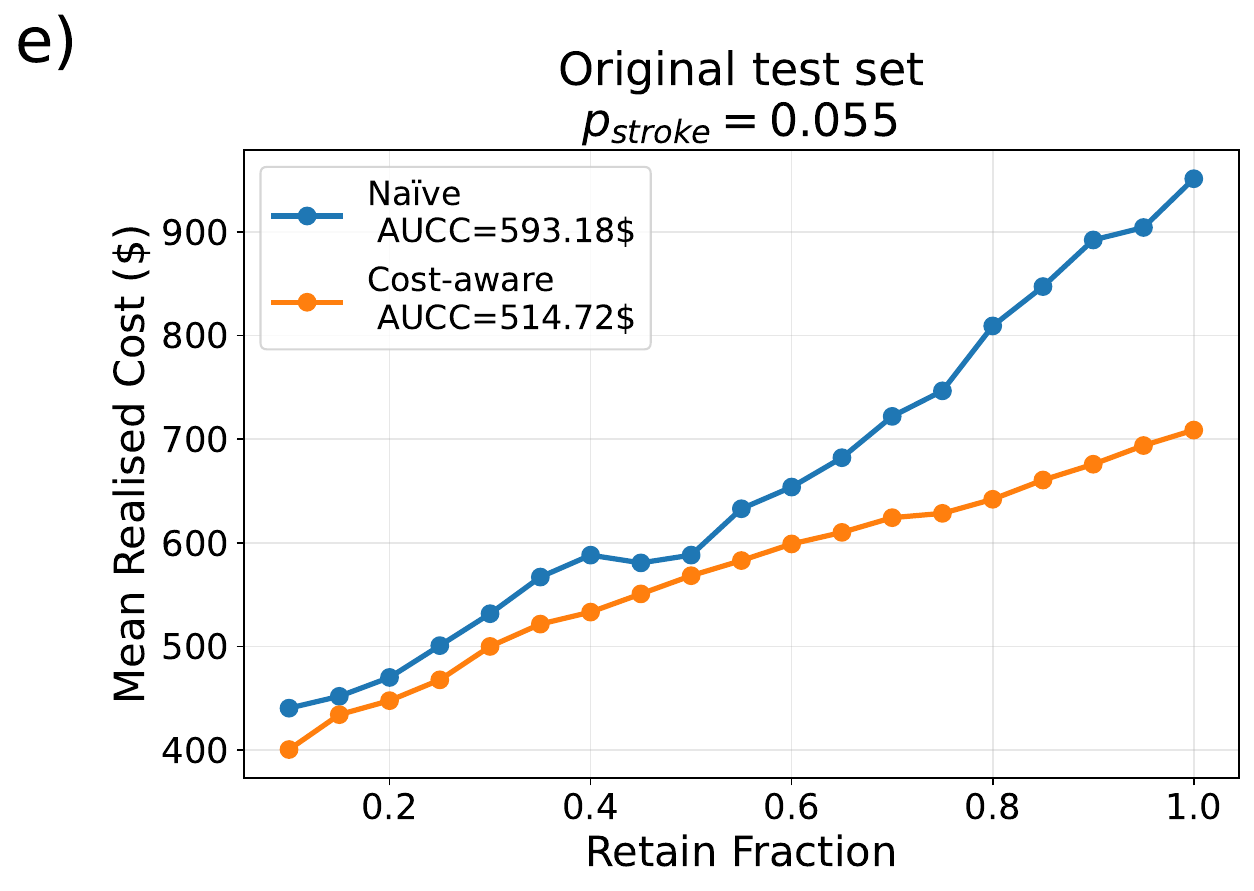}\includegraphics[width=0.45\linewidth]{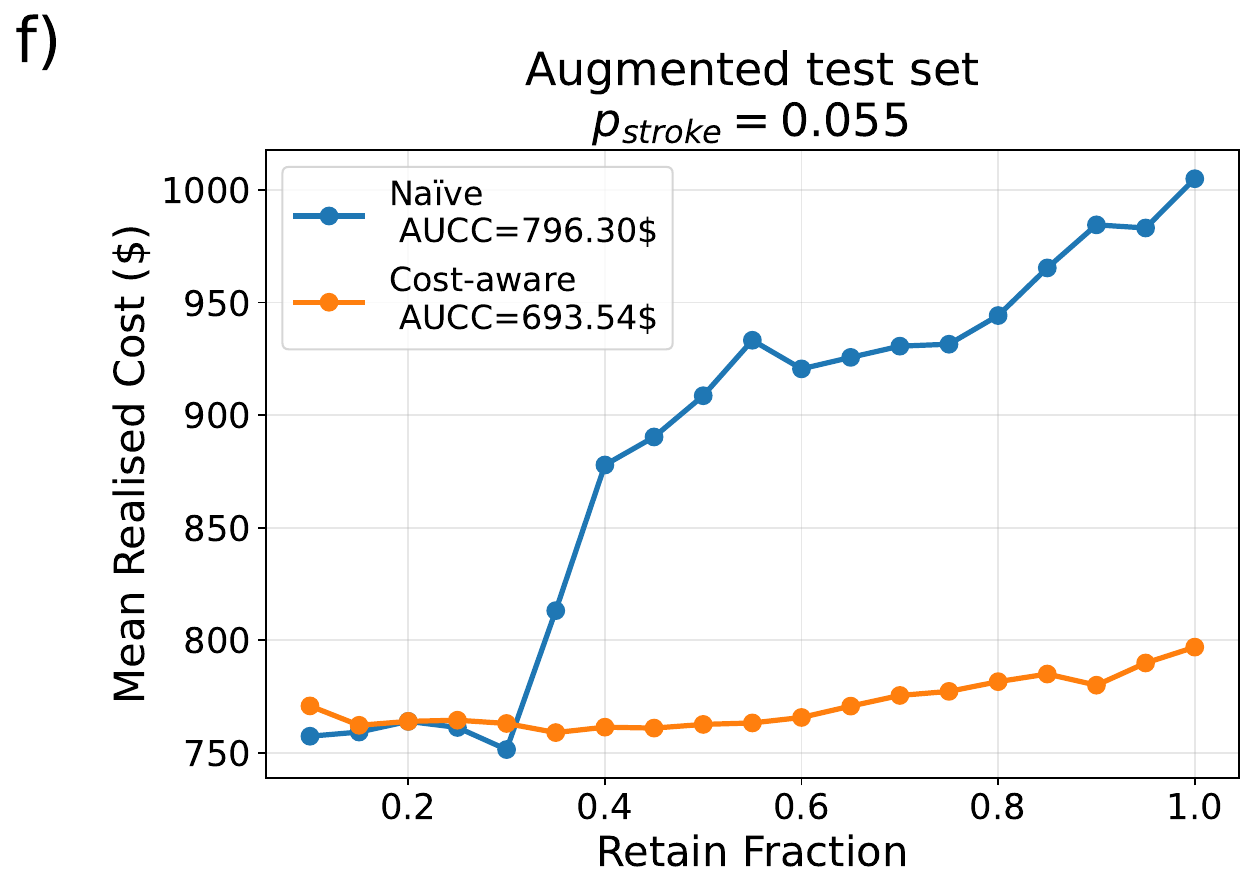}
    \caption{Cost-coverage curves and AUCC for the original and augmented test sets (with decision-optimal values of $\tau$ for the cost-aware rule, and $\tau=0.5$ for the na\"{i}ve rule).}
    \label{fig:coveragecurvesnoise03}
\end{figure}

\subsection{Decision-cost based evaluation}
We aim to evaluate whether cost-aware filtering rules based on decision-theoretic expressions of uncertainty result in lower realised cost among retained predictions compared to a more generic ranking of vanilla confidence scores. The model is evaluated on the unaugmented, and noise-augmented variants of the test set. We plot a cost-coverage curve for both sets for a range of prediction retain fractions (see Fig. \ref{fig:coveragecurvesnoise03}), where the cost is computed using the full cost model (Equation \ref{eq:realcost}). We compute the area under the cost-curves (AUCC) as an aggregate cost metric. Lower AUCC indicates that, across the evaluated retention levels, the filtering rule preferentially retains predictions associated with lower realised downstream cost. The Appendix includes results for the other noise augmentation strengths and decision thresholds and bootstrapped intervals (Appendices \ref{sec:supplement} and  \ref{sec:bootstrap}).

On the original test set, the decision-theoretic cost-aware filtering produced a lower AUCC for all three cost models. Relative to na\"ive filtering, AUCC decreased by 3.5\% for $p_{\mathrm{stroke}} = 0.025$, 8.1\% for $p_{\mathrm{stroke}} = 0.037$, and 13.2\% for $p_{\mathrm{stroke}} = 0.055$. The increasing improvement indicates that the benefit of cost-aware filtering became more pronounced as the asymmetry between the costs of different prediction outcomes increased.

Under the $\sigma=0.3$ noise-induced domain shift, the relative performance of the filtering rules again depended on the degree of cost asymmetry. For $p_{\mathrm{stroke}} = 0.025$, the cost-aware AUCC has a small increase of 0.5\%. For $p_{\mathrm{stroke}} = 0.037$, cost-aware filtering reduced AUCC by 3.8\%. For $p_{\mathrm{stroke}} = 0.055$, the reduction increased to 12.9\%. Though, greater differences in costs are observed for higher retain fractions, where cost-aware filtering appears to have a greater effect on reducing costs. For the na\"ive rule, the composition of examples (e.g. TP, FP, TN, FN rates) associated with each retain fraction is fixed, but FN costs are scaled with $p_\text{stroke}$ which can result in significant changes to the cost coverage curve.

The composition of the rejected subsets in Table \ref{tab:error-compositionnoise03} shows that the two filtering rules remove qualitatively different examples. The cost-aware filtering rule filters more false positives than the na\"{i}ve rule. A similar pattern is observed under OOD shift. The OOD results further show that distribution shift changes the error composition of the model, with false positives increasing substantially in the unfiltered OOD set. Appendix \ref{sec:tau5retain} shows that the asymmetry-dependent pattern persists with the use of a common threshold of $\tau=0.5$, but with smaller differences in AUCC. Broadly, the effectiveness of a rejection rule therefore depends on the types and frequencies of errors that occur after deployment. A filtering strategy designed around one error distribution may not be effective when domain shift changes the relative frequency of high-cost outcomes. 

The severity of the domain shift also influences cost differences. At $\sigma=0.1$, cost-aware filtering is beneficial for higher retain fractions when using the cost-aware threshold, but produces higher costs for lower retain fractions. When using $\tau=0.5$, there is little difference between the two filtering rules for higher retain fractions, but poor performance for cost-aware filtering at low retain fractions. At $\sigma=0.6$ the cost-aware rule provides benefit only at high retain fractions when there is high cost asymmetry. So while we find there is generally some benefit to using cost-aware filtering for higher retain fractions, model generalisability influences the utility of the filtering approaches.

\subsection{Limitations}
\label{sec:lim}
It's important to note that lower realised economic cost does not necessarily indicate improved patient outcomes \cite{fu2023considering}. Additionally, other works have discussed how prediction filtering may not provide performance benefits \cite{jabbour2025limits}.  While convenient for this analysis, the static form of the cost function may not accurately model decision costs in realistic clinically grounded deployment settings. For example, a single false-negative does not necessarily represent missed diagnosis, since AF missed in one window may be detected in a subsequent window within the same monitoring period. Applying an annual stroke-related cost independently to every false-negative window may not reflect the potential consequence accurately. As clarified previously, the cost model used here should be interpreted as a stylised/toy representation of asymmetric error consequences, rather than as a comprehensive and realistic estimate of economic cost. A clinically grounded cost model would instead apply the stroke-related consequence only when AF remains undetected across repeat measurements over some monitoring period; however, the length of streams in the test data prevent an effective analysis of this form. Also, these results are not guaranteed to generalise to other forms of calibration degradation that could emerge from other types of dataset shift. Nonetheless, this work demonstrates how decision-theoretic uncertainty may provide useful insights into how the utility of uncertainty estimates can vary depending on whether decision-cost is considered in the formulation of decision rules.
\subsection{Implications for communicating uncertainty} \label{sec:implications}
These results support a distinction between communicating predictive doubt and decision consequences. A confidence score may be interpreted as a heuristic indicator of prediction accuracy. It does not by itself indicate whether acting on the corresponding prediction would be beneficial. Expected decision cost considers the downstream decision made from the confidence score and may therefore provide a more decision-relevant description of uncertainty. The same prediction may correspond to different expected-cost values under different clinical pathways, patient populations, economic assumptions, or stakeholder priorities. Communicating a cost estimate without also communicating the associated decision model could create an unjustified impression of objectivity or generalisability. While here we have shown that confidence scores can be expressed through decision cost and may improve realised cost when used to formulate a prediction filtering strategy, whether this expression of uncertainty improves clinician comprehension, trust, or decision quality requires direct evaluation with intended users.

\section{Conclusions}
Here we have investigated the use of a decision-theoretic interpretation of uncertainty quantification to characterise Venn-Abers uncertainty reliability under the common failure mode of dataset shift (the noise augmentation considered here). Cost-aware filtering strategies provided lower decision cost on the original dataset relative to the na\"{i}ve strategy when the cost model had greater asymmetry across positive and negative predictions. However, the benefit in the OOD setting depended on both the severity of the signal corruption and cost asymmetry, with the greatest benefit occurring for larger retain fractions. While more generic calibration curves showed evidence of degradation in confidence calibration under dataset shift, these findings indicate that their utility also depends on the downstream decision model, the distribution of errors encountered after deployment, and how predictions are selected for use. 
\section*{Code}
The software and instructions for acquiring the datasets for this study are given here: \url{https://github.com/ciaranbench-npl/dtuqaf}

\bibliographystyle{unsrtnat}
\bibliography{bib.bib}
\appendix
\section{Cost model terms and screening pathways}
\label{app:costterms}
 The screening term is given as,
\[
C_{\mathrm{screen}} = C_{\mathrm{device}}  + C_{\mathrm{nurse}},
\]
where $C_{\mathrm{nurse}}$ is the cost of an initial assessment (\$76.15\cite{aasm2020em}), and  $C_{\mathrm{device}}$ is the cost of the wearable monitoring device (\$300 \cite{chen2022cost}).
$C_{\mathrm{confirm}}$ is the cost of a confirmatory patch-based testing (\$614 \cite{reynolds2023cost}), $C_{\mathrm{phys}}$ is the cost of a physician's follow-up assessment (\$76.15 \cite{aasm2020em}).  The breakdown of the cost models for the rest of the terms are given in the following sections. \textbf{We consider three variants of the cost model, each with a different cost associated with false negatives based on the probability of stroke occurring in the patient population}. 
\subsection{True positive}
For the treatment workflow/economic model for wearable devices described in \cite{chen2022cost}, a true positive incurs confirmatory testing, physician review, oral anticoagulation, a follow-up, and treatment-related bleeding cost, such that the cost may be modelled as,
\[
C_{\mathrm{TP}} = C_{\mathrm{confirm}} + C_{\mathrm{phys}} + C_{\mathrm{OAC}} + C_{\mathrm{follow}} + C_{\mathrm{bleed}}.
\]
\textbf{However, we set }$C_{\mathrm{TP}} = 0$ \textbf{to understand the \textit{net} consequences of incorrect decisions made from the use of predictions from wearable devices as opposed to estimating the gross-costs of deploying the model.}

$C_{\text{OAC}}$ is the cost of the oral anticoagulation treatment (\$2211.90 \cite{lamy2025cost}), $C_{\text{bleed}}$ is the cost of bleeding related treatment, where $C_{\text{bleed}} = (p_{bleed|AF})C_{bleed event}$, where $p_{bleed|AF} = 0.029$ is the annual major hemorrhage probability on OAC \cite{chen2022cost} and $C_{bleed event} = \$39236.33$ \cite{fermann2020major}, and $C_{follow}$ is the cost of the follow up visit, where $C_{follow} = C_{Phys}$.
\subsection{False positive}
A false positive incurs confirmatory testing and physician review but no further treatment, such that
\[
C_{\mathrm{FP}} = C_{\mathrm{confirm}} + C_{\mathrm{phys}}.
\]
\subsection{False negative}
A false negative is assigned the downstream cost of missed AF-related stroke,
\[
C_{\mathrm{FN}} = C_{\mathrm{stroke}}.
\]

where $C_{\mathrm{stroke}}$ is the expected cost of stroke-related harm associated with missed AF, and is given as $C_{\mathrm{stroke}} = (p_{\text{stroke|missed AF}}) C_{\text{stroke event}}$, where $p_{\text{stroke|missed AF}}$ is the conditional probability of a stroke occurring given a missed diagnosis. \cite{chen2022cost} does not assume a single fixed probability of stroke per false negative; instead, their model uses CHA$_2$DS$_2$-VASc--specific annual stroke incidence rates for AF without treatment, together with treatment-specific relative risks for anticoagulation, to model stroke risk over time. However, here, we employ a single probability to facilitate estimation of decision cost in light of missing patient metadata needed for these more complex models. Therefore, this is estimated as $p_{\text{stroke|missed AF}} = 0.025$ which corresponds to CHA$_2$DS$_2$-VASc of 2, 2.5\%/year, which is a reasonable lower-end OAC-eligible AF risk in older adults \cite{chen2022cost}. A cost of \$44,929 for treatment is given in \cite{johnson2016short}. We also consider  $p_{\text{stroke|missed AF}} = 0.037$ which corresponds to CHA$_2$DS$_2$-VASc of 3, 3.7\%/year, and $p_{\text{stroke|missed AF}} = 0.055$ which corresponds to CHA$_2$DS$_2$-VASc of 4, 5.5\%/year.

\subsection{True negative}
The true-negative state incurs zero cost,
\[
C_{\mathrm{TN}} = 0.
\]

\begin{table}[htbp]
\centering
\scriptsize
\setlength{\tabcolsep}{4pt}
\renewcommand{\arraystretch}{1.25}
\begin{tabular}{p{2.2cm} p{1.5cm} p{1.9cm} p{4.8cm} p{2.6cm}}
\hline
\textbf{Quantity} & \textbf{Symbol} & \textbf{Base-case value} & \textbf{Interpretation / use in reduced model} & \textbf{Reference} \\
\hline

AF prevalence & $\pi$ & Estimated from test set & Base-case AF prevalence for adults aged $\geq 65$ years; used to weight confusion-matrix terms in the reduced expected-cost expression. & n/a\\

Sensitivity & $\mathrm{Se}$ & Estimated from test set & Sensitivity of wearable photoplethysmography (PPG), used in $\pi \mathrm{Se}$ and $\pi(1-\mathrm{Se})$. & n/a \\

Specificity & $\mathrm{Sp}$ & Estimated from test set & Specificity of wearable photoplethysmography (PPG), used in $(1-\pi)(1-\mathrm{Sp})$ and $(1-\pi)\mathrm{Sp}$. & n/a \\

TP weight & $\pi \mathrm{Se}$ & Estimated from test set & Coefficient multiplying $C_{\mathrm{TP}}$ in the reduced expected-cost equation. & n/a \\

FN weight & $\pi(1-\mathrm{Se})$ & Estimated from test set & Coefficient multiplying $C_{\mathrm{FN}}$ in the reduced expected-cost equation. &  n/a \\

FP weight & $(1-\pi)(1-\mathrm{Sp})$ & Estimated from test set & Coefficient multiplying $C_{\mathrm{FP}}$ in the reduced expected-cost equation. & n/a \\

TN weight & $(1-\pi)\mathrm{Sp}$ & Estimated from test set & Coefficient multiplying $C_{\mathrm{TN}}$ in the reduced expected-cost equation. & n/a\\

Screening nurse visit proxy & $C_{\mathrm{nurse}}$ & \$76.15 & Proxy one-time nurse/administration cost at screening. &  \cite{aasm2020em} \\

Wearable device cost scenarios & $C_{\mathrm{device}}$ & \$300 & Device-cost scenarios explicitly evaluated by the wearable AF screening model; may be included in $C_{\mathrm{screen}}$ if desired. & \cite{chen2022cost} \\

Screening cost (with device) & $C_{\mathrm{screen}}$ & \$376.15 & Device scenario plus one screening nurse-visit proxy: $C_{\mathrm{device}} + C_{\mathrm{nurse}}$. & \cite{chen2022cost,aasm2020em} \\

Confirmatory monitoring proxy & $C_{\mathrm{confirm}}$ & \$614.00 \cite{reynolds2023cost} & External US proxy for confirmatory 2-week patch monitor after a positive wearable signal. & \cite{reynolds2023cost} \\

Physician visit proxy & $C_{\mathrm{phys}}$ & \$76.15 & Proxy physician cost for diagnosis counselling / follow-up. & \cite{aasm2020em} \\

False-positive cost & $C_{\mathrm{FP}}$ & \$690.15 & Reduced-form false-positive pathway cost: $C_{\mathrm{confirm}} + C_{\mathrm{phys}}$. & Computed as \$614.00 \cite{reynolds2023cost} + \$76.15 \cite{aasm2020em}; pathway structure supported by \cite{chen2022cost}. \\

OAC cost proxy & $C_{\mathrm{OAC}}$ & \$2211.90/year  & Annual anticoagulation cost proxy based on US apixaban daily cost (\$6.06/day). &  \cite{lamy2025cost}  \\

Follow-up visit proxy & $C_{\mathrm{follow}}$ & \$76.15/year & Annual follow-up physician visit proxy for anticoagulated patients. & \cite{aasm2020em} \\

Major bleed incidence on DOAC & $p_{\mathrm{bleed}}$ & 0.029/year & Annual major hemorrhage probability on DOAC. & \cite{chen2022cost} \\

Mean major bleed event cost proxy & --- & \$39{,}236.33 & Simple mean of reported major bleed hospitalisation costs across GI bleed, intracranial haemorrhage, and other major bleed. & \cite{fermann2020major}\\

Expected bleed cost & $C_{\mathrm{bleed}}$ & \$1137.85/year & Expected annual bleed cost used in the base-case $C_{\mathrm{TP}}$, computed as $0.029 \times 39{,}236.33$. & Computed from DOAC major hemorrhage incidence in \cite{chen2022cost} supplement and mean bleed event cost from \cite{fermann2020major}. \\

True-positive cost & $C_{\mathrm{TP}}$ & \$0  & Set to \$0 to enable modelling of net costs. \\

False-negative cost & $C_{\mathrm{FN}}$ & \$1123.23, \$1662.37, \$2471.10 & One-year acute ischemic stroke cost surrogate used to represent downstream harm from missed AF in the reduced model. & Computed as \$44{,}929 ($p_{\text{stroke|missed AF}} = 0.025$) where $p_{\text{stroke|missed AF}}$ corresponds to CHA$_2$DS$_2$-VASc of 2, 2.5\%/year \cite{chen2022cost}\cite{johnson2016short}. We also consider  $p_{\text{stroke|missed AF}} = 0.037$ which corresponds to CHA$_2$DS$_2$-VASc of 3, 3.7\%/year, and $p_{\text{stroke|missed AF}} = 0.055$ which corresponds to CHA$_2$DS$_2$-VASc of 4, 5.5\%/year. \\

True-negative cost & $C_{\mathrm{TN}}$ & \$0 & Reference state in the reduced model; all fixed screening expenditure is already captured in $C_{\mathrm{screen}}$. & Modelling assumption for reduced-form deployment-cost analysis. \\

\hline
\end{tabular}
\caption{Summary of the base-case parameterisation used for the reduced-form AF classification cost model.}
\label{tab:basic_cost_parameters}
\end{table}

\newpage
\section{Derivation of the realised cost model}
\label{app:derivrealcost}
Here we provide the derivation for the realised cost model in Equation \ref{eq:fullrealcost}. Let $Y \in \{0,1\}$ denote the true patient state, where $Y=1$ indicates atrial fibrillation (AF), and let $\hat{Y} \in \{0,1\}$ denote the model prediction, where $\hat{Y}=1$ indicates a positive AF prediction. Let the prevalence of AF be,
\[
P(Y=1)=\pi,
\]
so that,
\[
P(Y=0)=1-\pi.
\]
Sensitivity and specificity are defined as,
\[
\mathrm{Se}=P(\hat{Y}=1 \mid Y=1),
\]
and
\[
\mathrm{Sp}=P(\hat{Y}=0 \mid Y=0).
\]
Therefore, the error probabilities are given by,
\[
P(\hat{Y}=0 \mid Y=1)=1-\mathrm{Se},
\]
and,
\[
P(\hat{Y}=1 \mid Y=0)=1-\mathrm{Sp}.
\]
The multiplication rule gives,
\[
P(Y,\hat{Y}) = P(\hat{Y}\mid Y)P(Y),
\]
so the probabilities of the four outcomes may be expressed as:
\begin{align}
P(\mathrm{TP})
&= P(Y=1,\hat{Y}=1)
= P(\hat{Y}=1\mid Y=1)P(Y=1)
= \pi \mathrm{Se}, \\
P(\mathrm{FN})
&= P(Y=1,\hat{Y}=0)
= P(\hat{Y}=0\mid Y=1)P(Y=1)
= \pi(1-\mathrm{Se}), \\
P(\mathrm{FP})
&= P(Y=0,\hat{Y}=1)
= P(\hat{Y}=1\mid Y=0)P(Y=0)
= (1-\pi)(1-\mathrm{Sp}), \\
P(\mathrm{TN})
&= P(Y=0,\hat{Y}=0)
= P(\hat{Y}=0\mid Y=0)P(Y=0)
= (1-\pi)\mathrm{Sp}.
\end{align}
The realised cost is therefore the probability-weighted sum of the costs associated with each diagnostic outcome:
\begin{align}
\mathbb{E}[C]
&=
P(\mathrm{TP})C_{\mathrm{TP}}
+
P(\mathrm{FN})C_{\mathrm{FN}}
+
P(\mathrm{FP})C_{\mathrm{FP}}
+
P(\mathrm{TN})C_{\mathrm{TN}} \\
&=
\pi\mathrm{Se}C_{\mathrm{TP}}
+
\pi(1-\mathrm{Se})C_{\mathrm{FN}}
+
(1-\pi)(1-\mathrm{Sp})C_{\mathrm{FP}}
+
(1-\pi)\mathrm{Sp}C_{\mathrm{TN}}.
\end{align}

Including a fixed screening cost \(C_{\mathrm{screen}}\) incurred for each
screened individual, cost is defined as,
\begin{align}
\bar C
&= C_{\text{screen}}
+ \pi\left[\mathrm{Se}\,C_{\text{TP}} + (1-\mathrm{Se})\,C_{\text{FN}}\right]
+ (1-\pi)\left[(1-\mathrm{Sp})\,C_{\text{FP}} + \mathrm{Sp}\,C_{\text{TN}}\right].
\end{align} 

\section{Decision-optimal threshold derivation}
\label{sec:derivdot}
The AF-positive pathway is selected when its expected cost is lower than 
that of the AF-negative pathway:
\begin{equation}
pC_{\mathrm{TP}} + (1-p)C_{\mathrm{FP}} \leq pC_{\mathrm{FN}} + (1-p)C_{\mathrm{TN}}.
\end{equation}
Solving this inequality for $p$ gives the cost-optimal decision threshold
\begin{equation}
\tau^{*} = \frac{C_{\mathrm{FP}}-C_{\mathrm{TN}}}
{\left(C_{\mathrm{FP}}-C_{\mathrm{TN}}\right)
+\left(C_{\mathrm{FN}}-C_{\mathrm{TP}}\right)}
\end{equation}
such that the AF-positive pathway is selected when \(p \geq \tau^{*}\), assuming \(C_{\mathrm{FN}} > C_{\mathrm{TP}}\).

\newpage

\newpage
\clearpage
\section{Supplemental experimental data}
\begin{figure}[h]
    \centering
    \includegraphics[width=.9\linewidth]{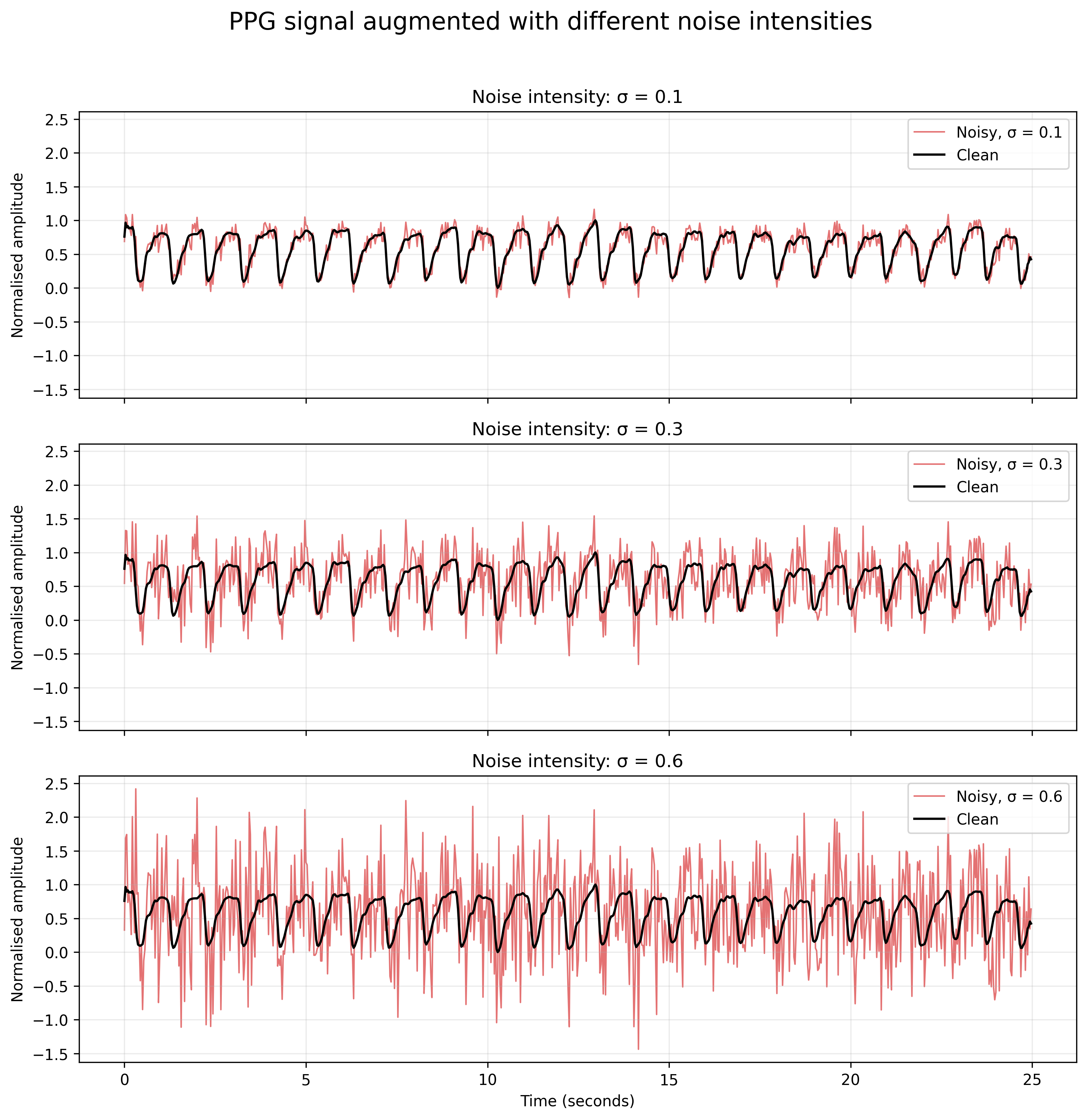}
    \caption{The same PPG signal augmented with noise parameterised by $\sigma$.}
    \label{fig:cleannoisyex}
\end{figure}
\label{sec:supplement}
\subsection{$\sigma=0.1$}
\subsubsection{Decision optimal threshold}
\begin{figure}[h]
    \centering
    Noise $\sigma=0.1$, Decision optimal $\tau$ for cost-aware rule
    \includegraphics[width=0.45\linewidth]{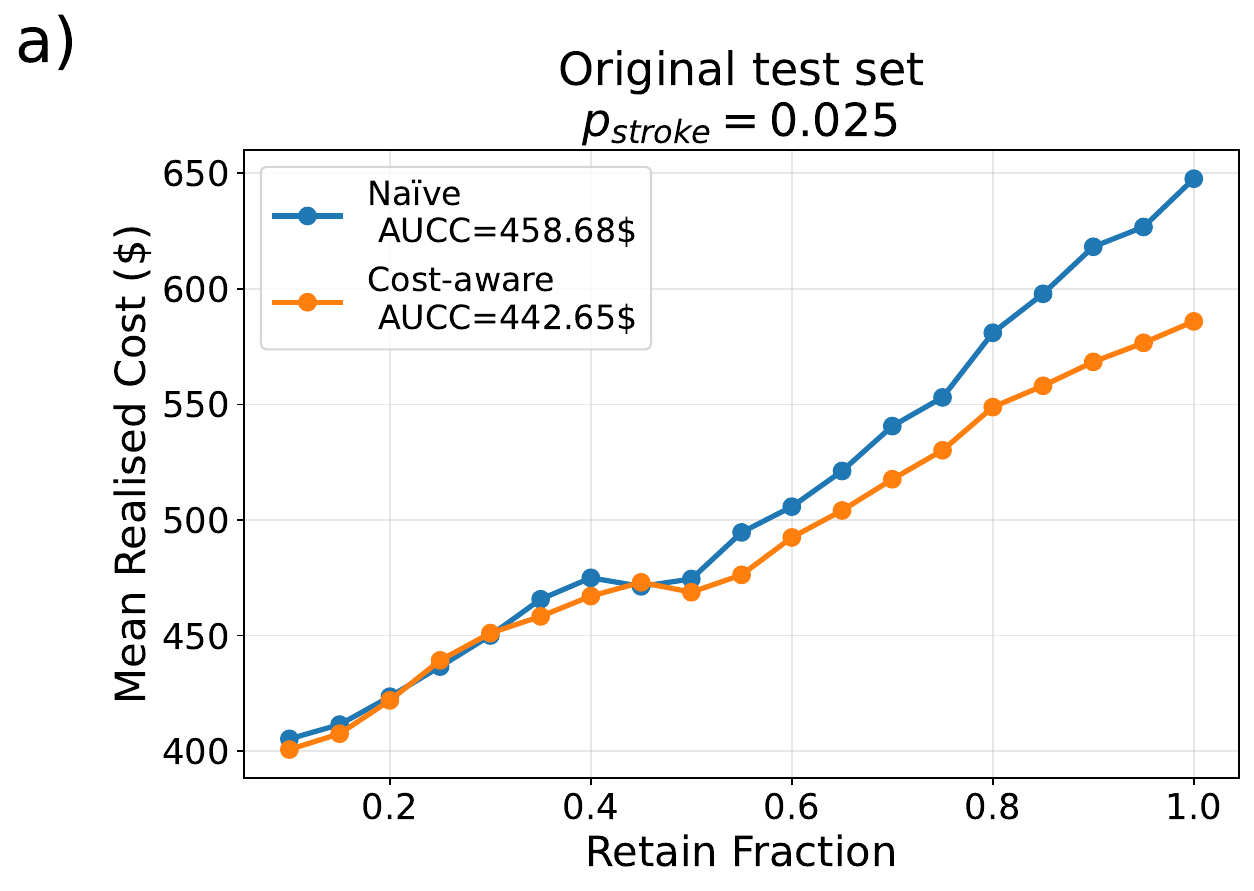}\includegraphics[width=0.45\linewidth]{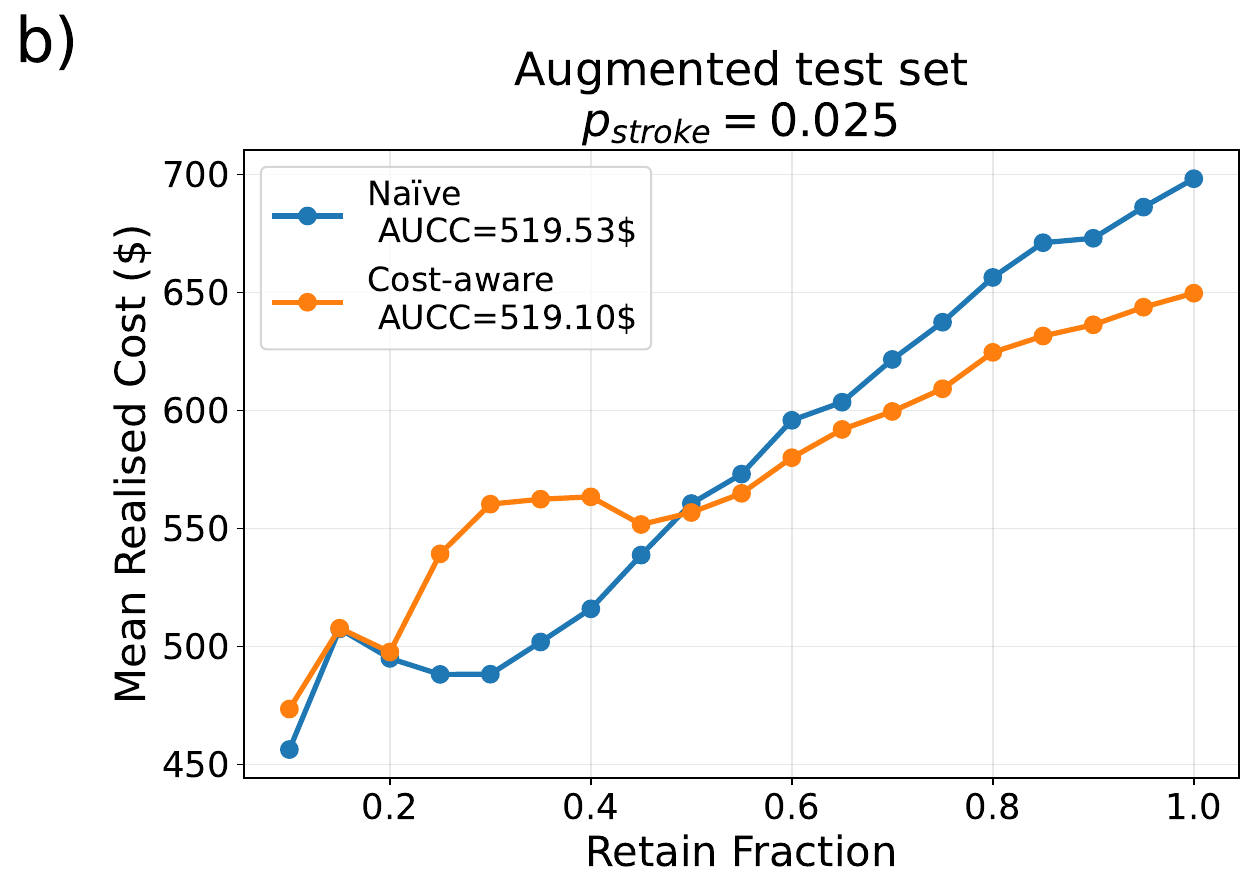}
    \includegraphics[width=0.45\linewidth]{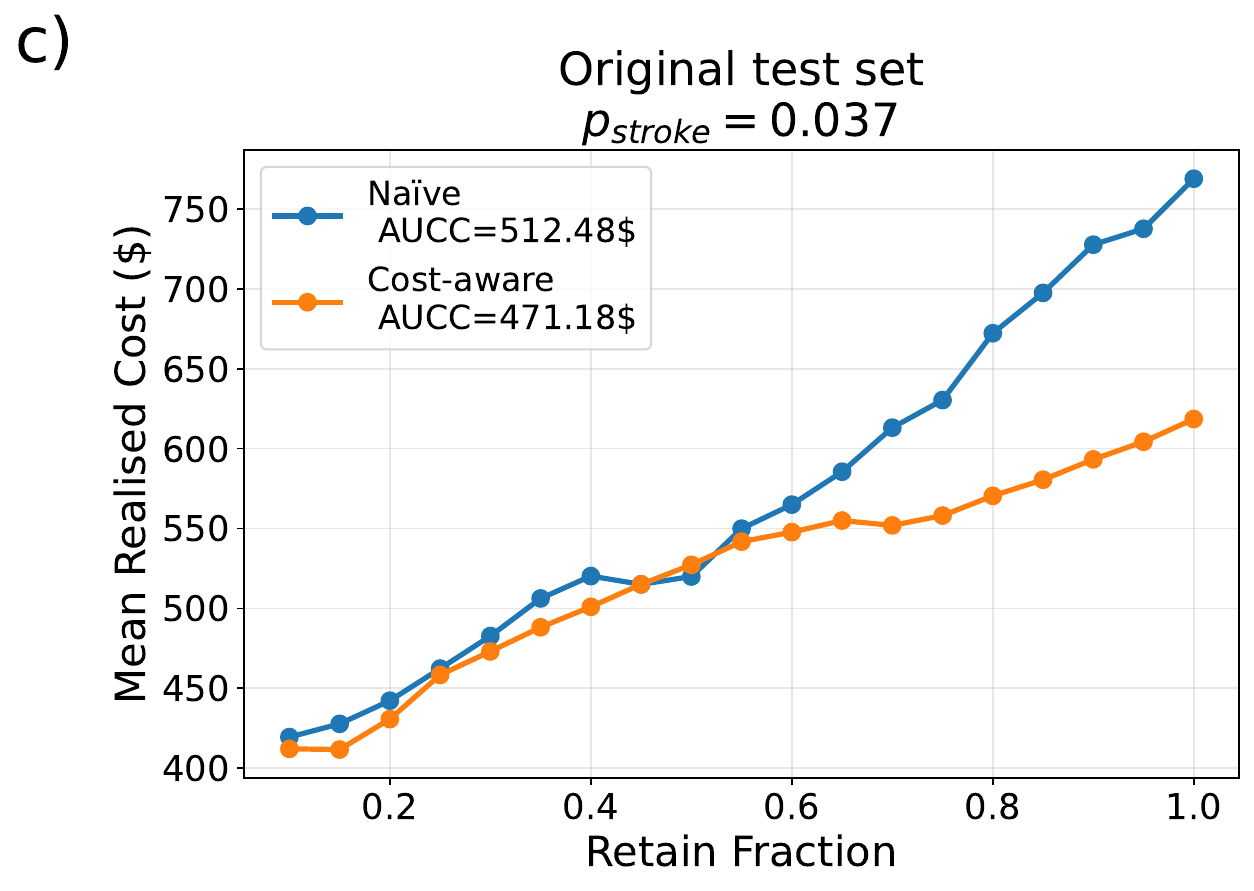}
    \includegraphics[width=0.45\linewidth]{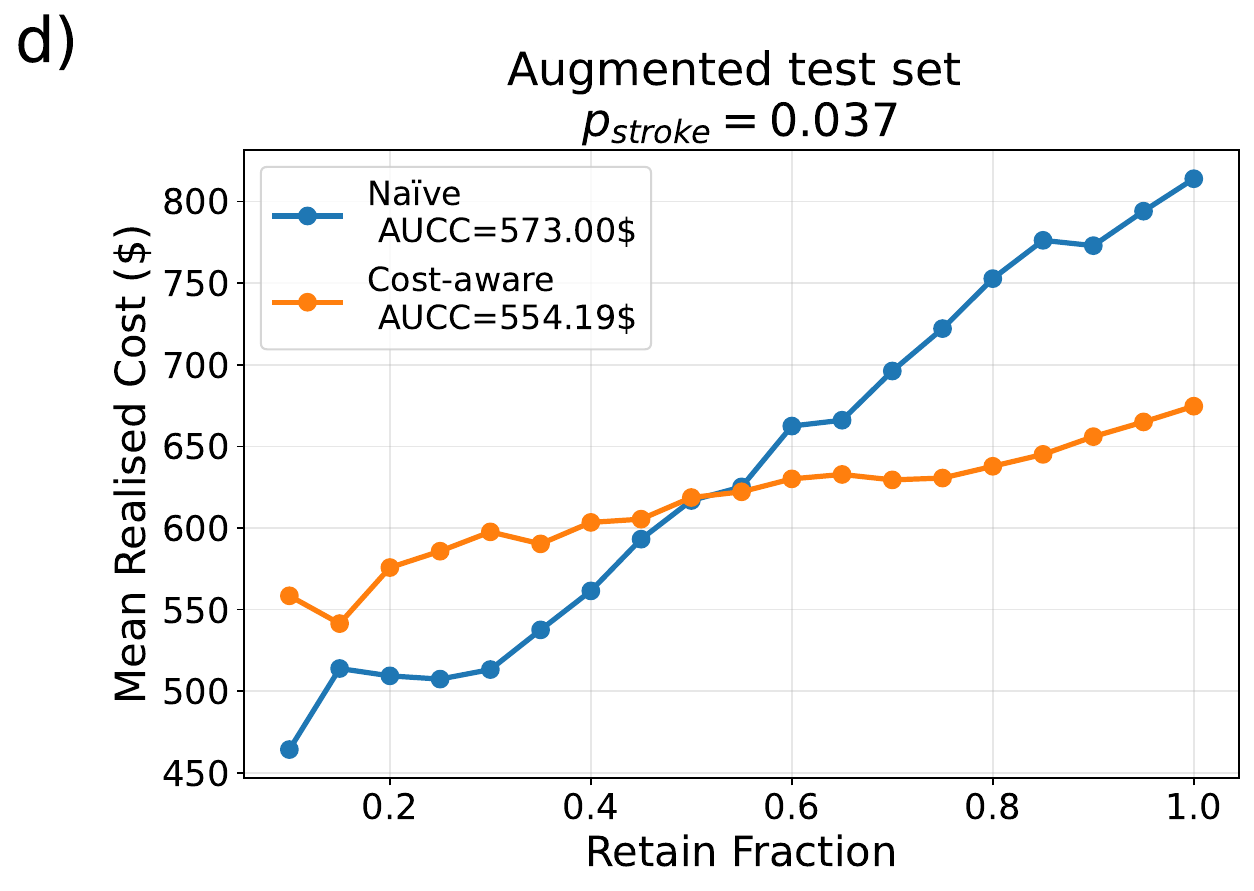}
    \includegraphics[width=0.45\linewidth]{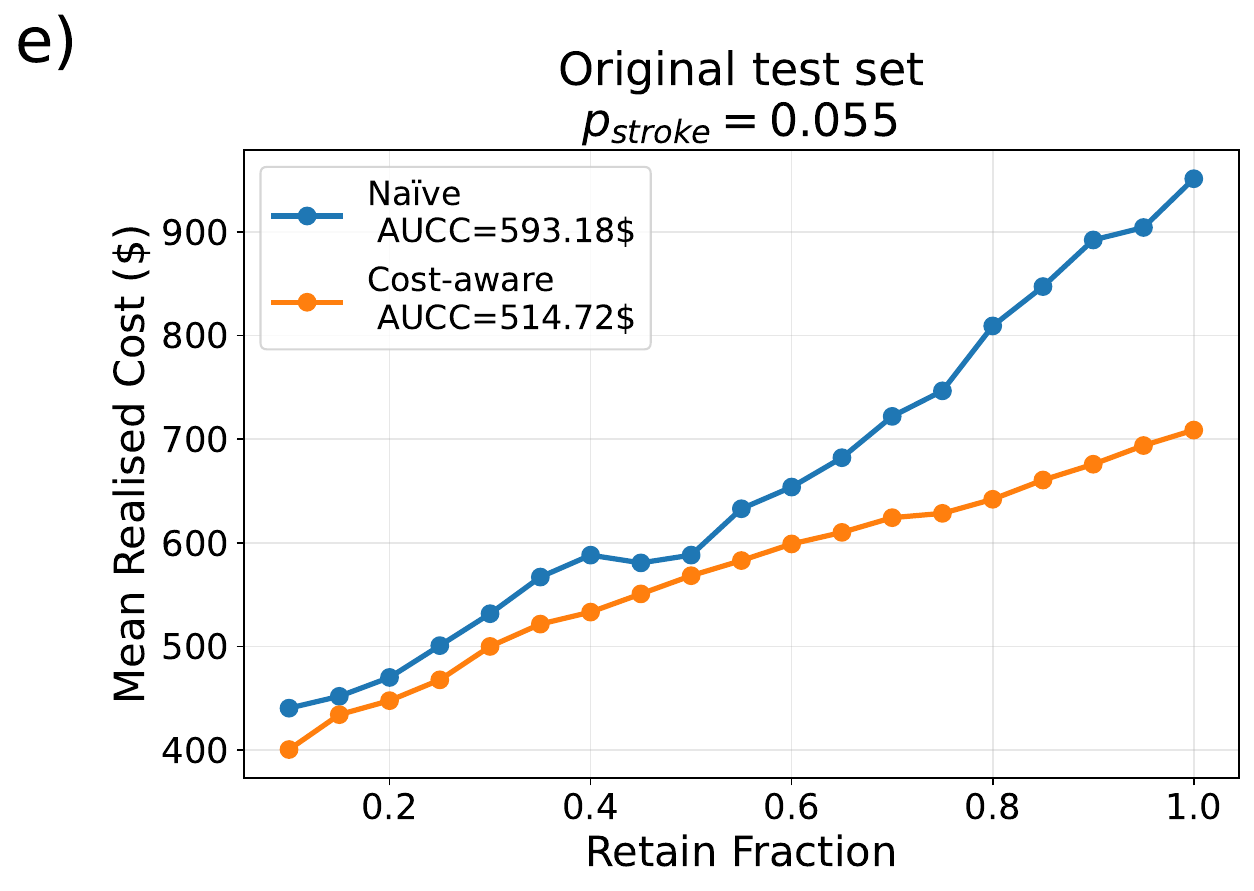}\includegraphics[width=0.45\linewidth]{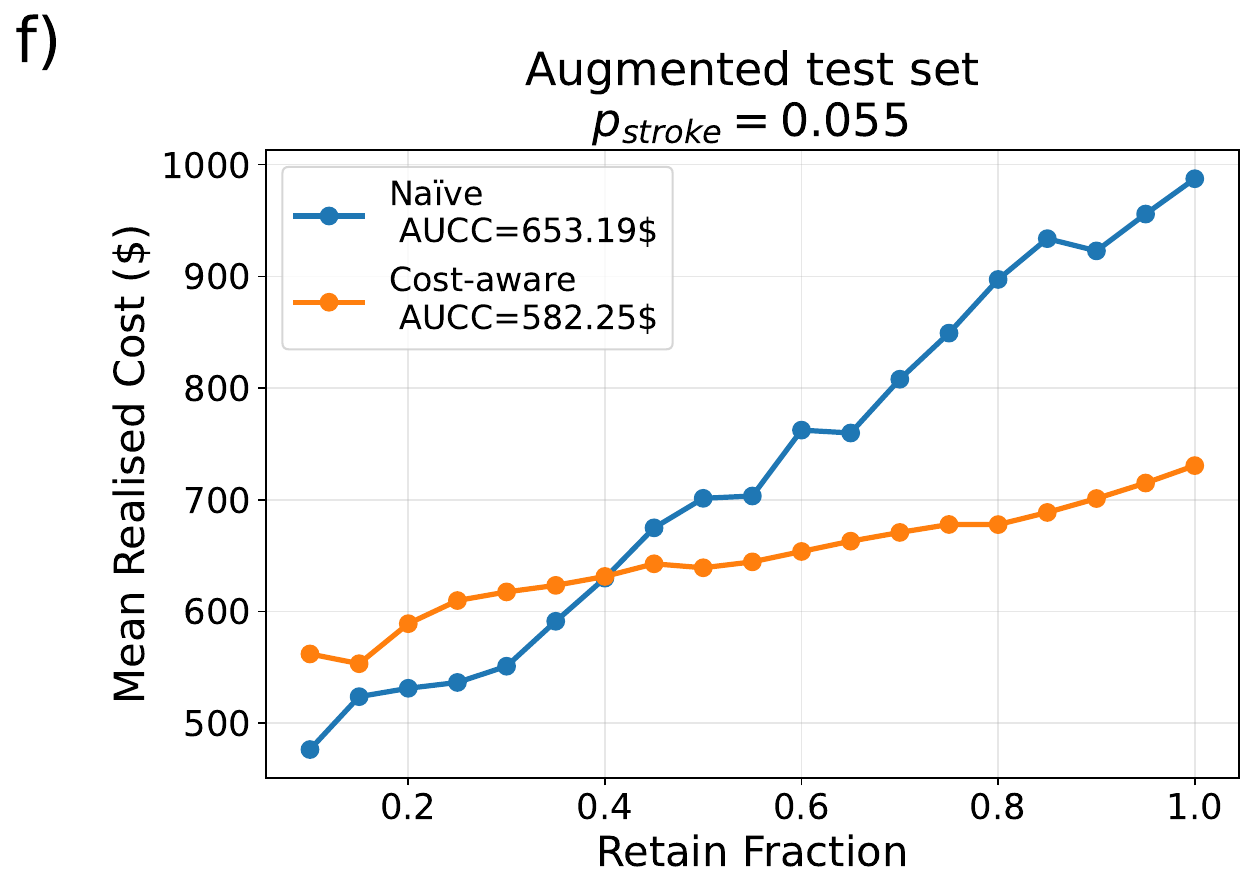}
    \caption{Cost-coverage curves and AUCC for the original and augmented test sets (with decision-optimal values of $\tau$ for the cost-aware rule, and $\tau=0.5$ for the na\"{i}ve rule).}
    \label{fig:coveragecurvesnoise-1}
\end{figure}

\begin{table}[h] \centering \scriptsize \caption{Composition of prediction outcomes for the overall test set and rejected subset at a 50\% retain fraction for $\sigma=0.1$. N denotes the na\"{i}ve rule and C denotes the cost-aware rule. Percentages are calculated relative to the total number of examples in the corresponding subset. We use decision-optimal values of $\tau$ for the cost-aware rule and $\tau=0.5$ for the na\"{i}ve rule.} \label{tab:error-compositionnoise01} \begin{tabular}{lllrrrrr} \toprule $p_{\mathrm{stroke}}$ & Rule/data & Subset & Total & TN, n (\%) & FN, n (\%) & TP, n (\%) & FP, n (\%) \\ \midrule All & N, augmented & Overall & 15377 & 7774 (50.56\%) & 3299 (21.45\%) & 2498 (16.25\%) & 1806 (11.74\%) \\ All & N, original & Overall & 15377 & 9171 (59.64\%) & 3464 (22.53\%) & 2333 (15.17\%) & 409 (2.66\%) \\ \midrule All & N, augmented & Rejected & 7689 & 2912 (37.87\%) & 2496 (32.46\%) & 1222 (15.89\%) & 1059 (13.77\%) \\ All & N, original & Rejected & 7689 & 3408 (44.32\%) & 2816 (36.62\%) & 1097 (14.27\%) & 368 (4.79\%) \\ \midrule 0.025 & C, augmented & Rejected & 7689 & 1529 (19.89\%) & 787 (10.24\%) & 2571 (33.44\%) & 2802 (36.44\%) \\ 0.025 & C, original & Rejected & 7689 & 2335 (30.37\%) & 975 (12.68\%) & 2325 (30.24\%) & 2054 (26.71\%) \\ \midrule 0.037 & C, augmented & Rejected & 7689 & 2761 (35.91\%) & 687 (8.93\%) & 1948 (25.33\%) & 2293 (29.82\%) \\ 0.037 & C, original & Rejected & 7689 & 3177 (41.32\%) & 694 (9.03\%) & 1772 (23.05\%) & 2046 (26.61\%) \\ \midrule 0.055 & C, augmented & Rejected & 7689 & 479 (6.23\%) & 46 (0.60\%) & 2361 (30.71\%) & 4803 (62.47\%) \\ 0.055 & C, original & Rejected & 7689 & 731 (9.51\%) & 98 (1.27\%) & 1941 (25.24\%) & 4919 (63.97\%) \\ \bottomrule \end{tabular} \end{table}

\begin{table}[h] \centering \small \caption{Area under the cost--coverage curve (AUCC) for the na\"{i}ve and cost-aware filtering rules on the original and noise-augmented test sets for $\sigma=0.1$. Lower AUCC indicates lower realised cost across the evaluated retain fractions. The na\"{i}ve rule uses $\tau=0.5$, whereas the cost-aware rule uses the decision-optimal thresholds $\tau=\{0.381,0.293,0.218\}$ for $p_{\mathrm{stroke}}=\{0.025,0.037,0.055\}$, respectively.} \label{tab:auccnoise01} \begin{tabular}{llrr} \toprule $p_{\mathrm{stroke}}$ & Test set & Na\"{i}ve AUCC (\$) & Cost-aware AUCC (\$) \\ \midrule 0.025 & Original & 458.6814 & 442.6533 \\ 0.025 & Augmented & 519.5305 & 519.1049 \\ \midrule 0.037 & Original & 512.4814 & 471.1761 \\ 0.037 & Augmented & 572.9958 & 554.1864 \\ \midrule 0.055 & Original & 593.1815 & 514.7212 \\ 0.055 & Augmented & 653.1937 & 582.2468 \\ \bottomrule \end{tabular} \end{table}

\clearpage
\subsubsection{$\tau=0.5$}

\begin{figure}[h]
    \centering
    Noise $\sigma=0.1$, $\tau=0.5$
    \includegraphics[width=0.45\linewidth]{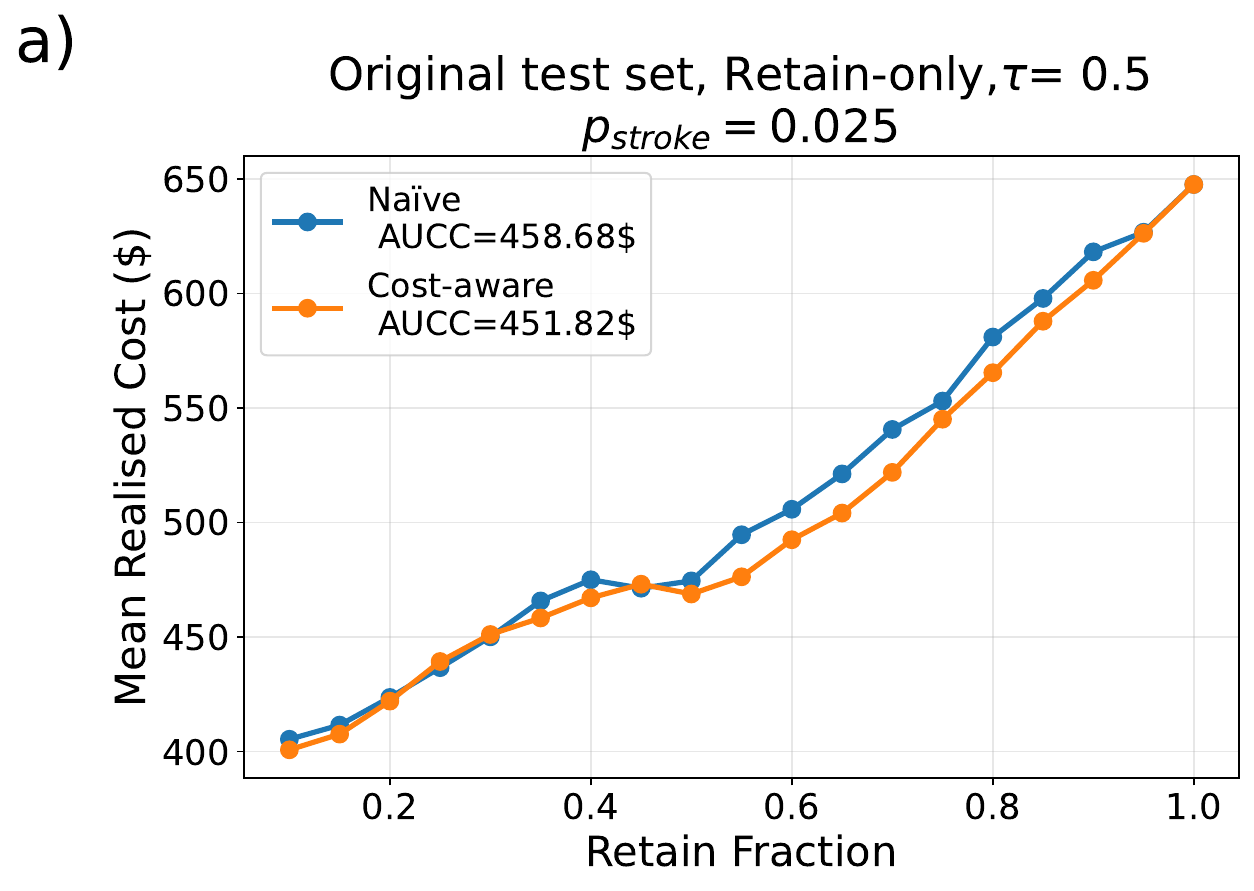}\includegraphics[width=0.45\linewidth]{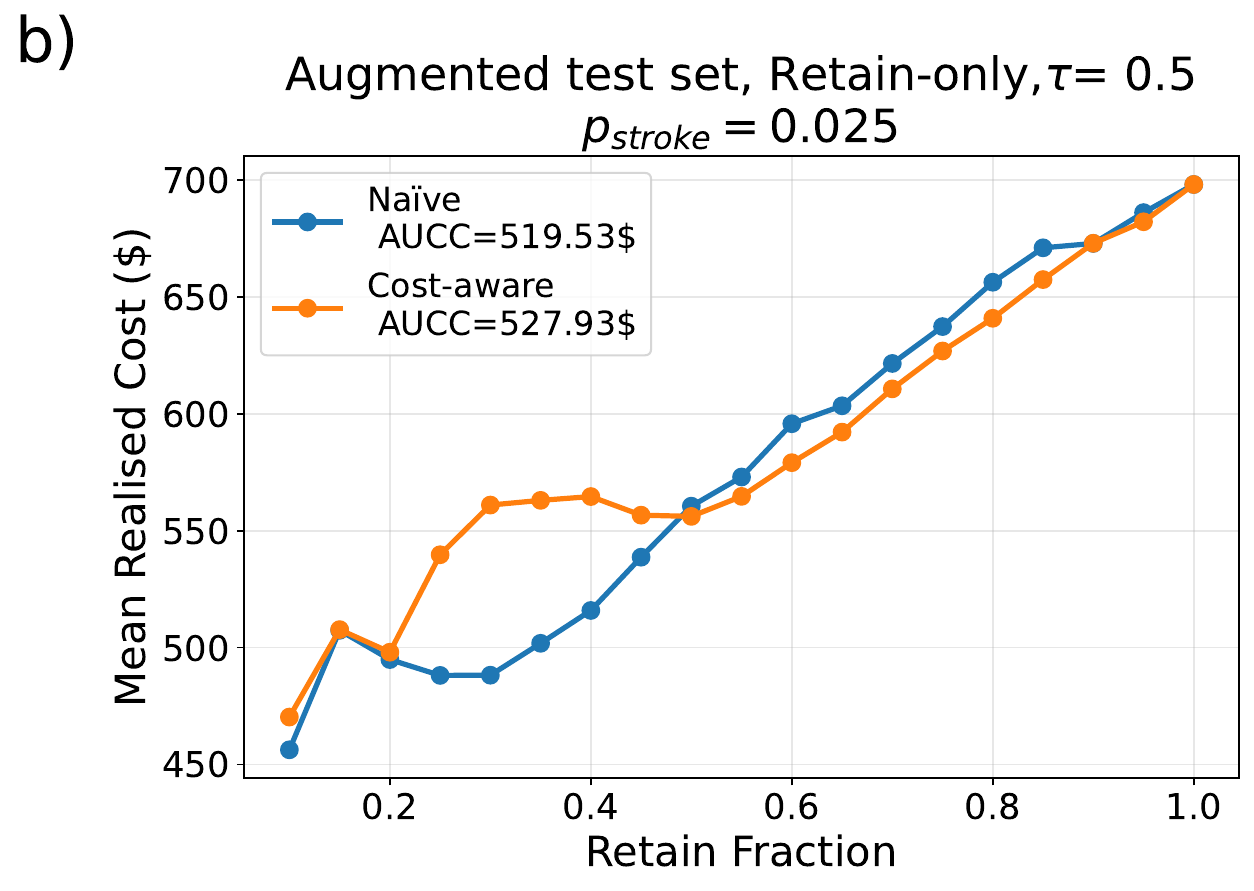}
    \includegraphics[width=0.45\linewidth]{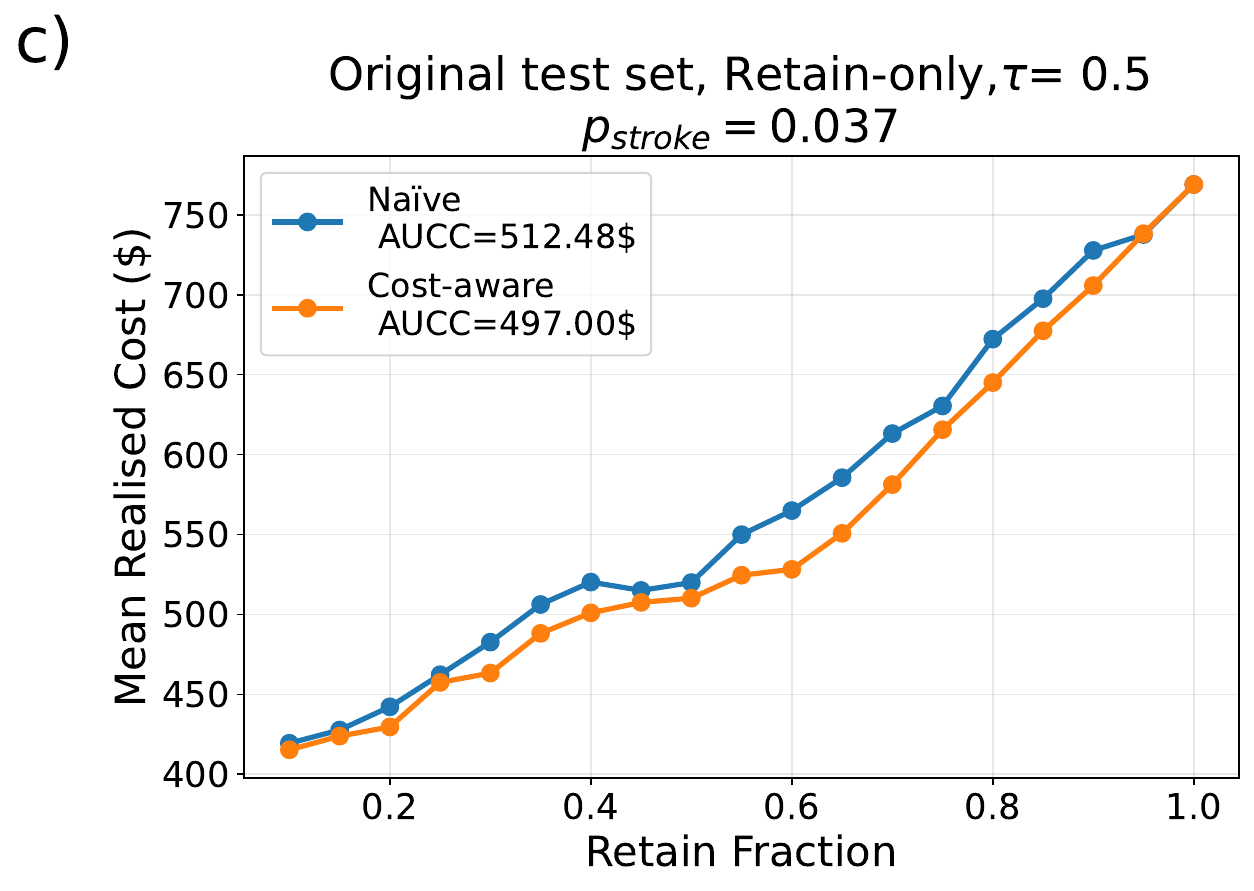}
    \includegraphics[width=0.45\linewidth]{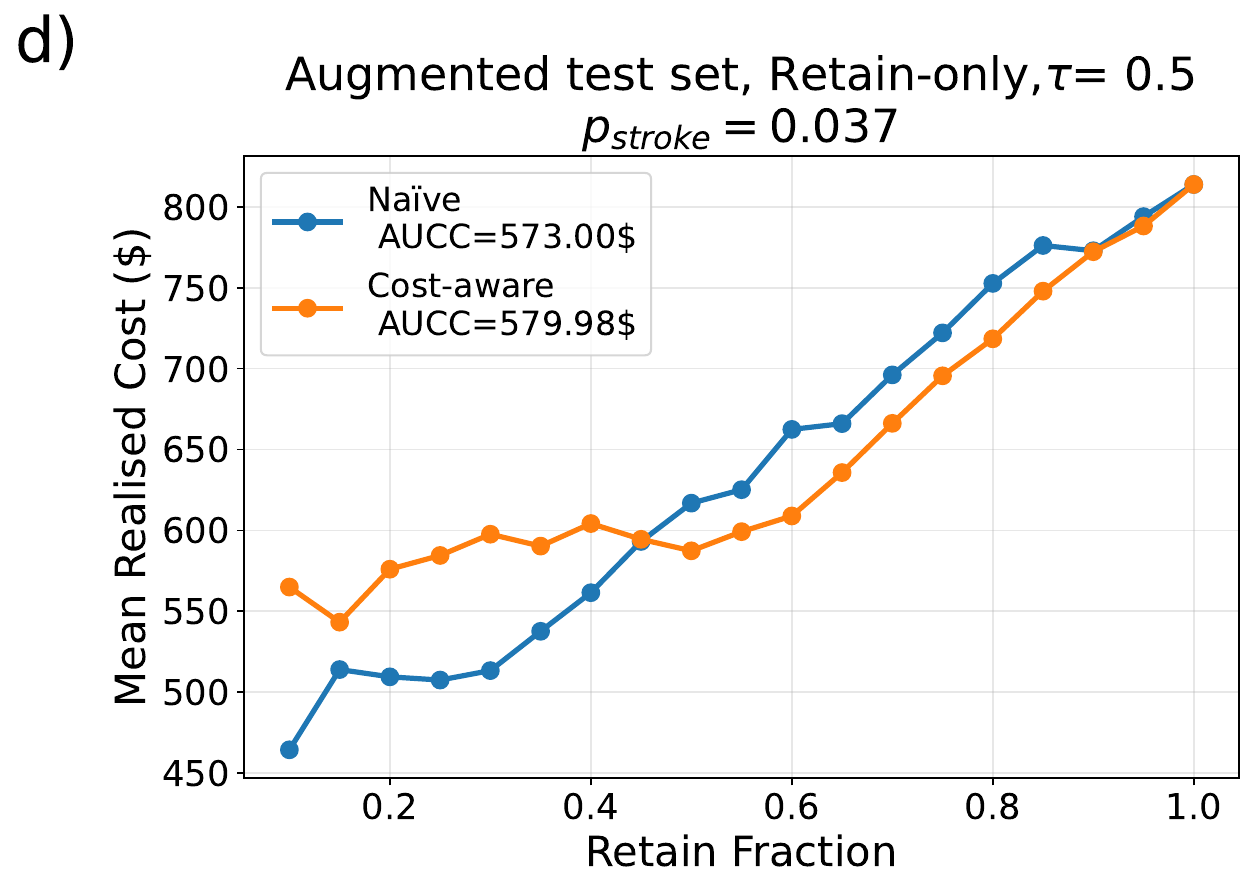}
    \includegraphics[width=0.45\linewidth]{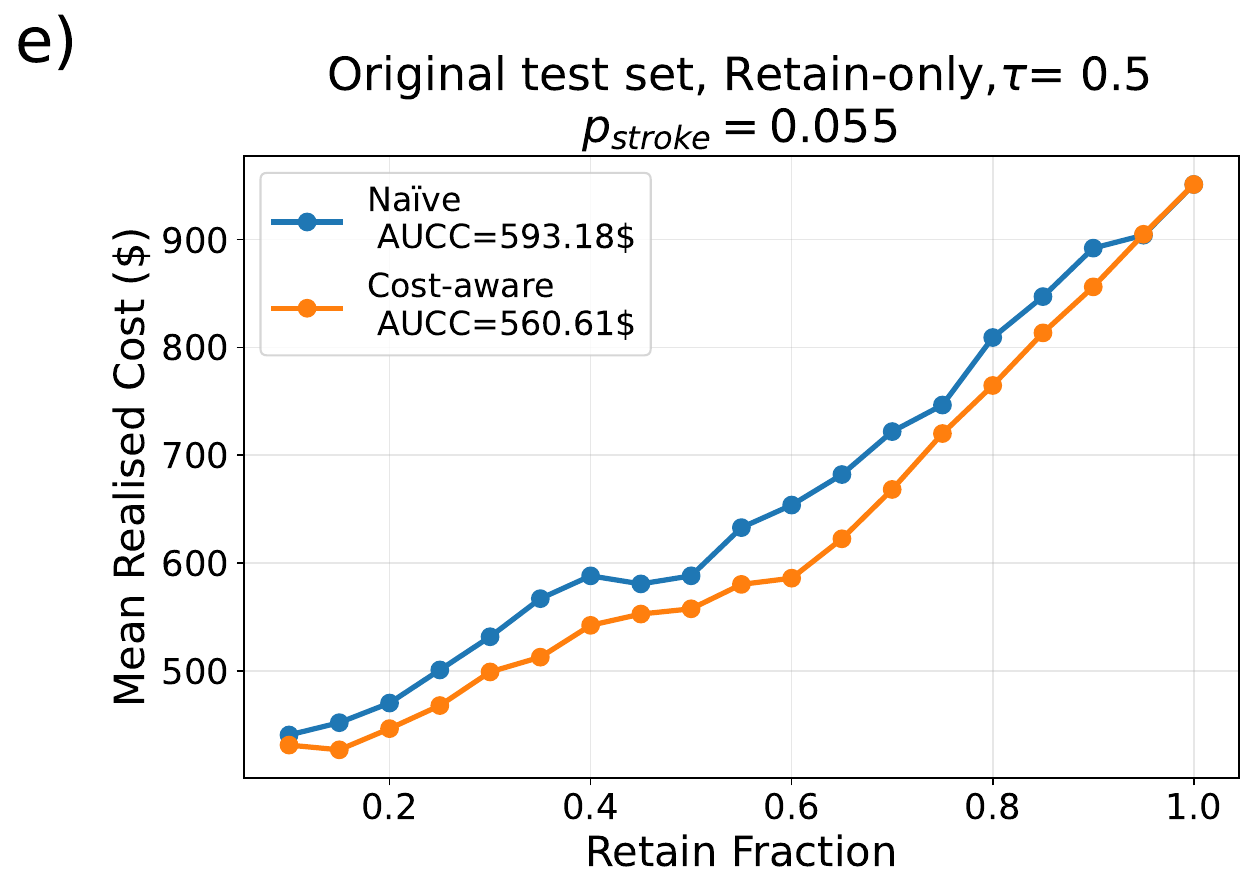}\includegraphics[width=0.45\linewidth]{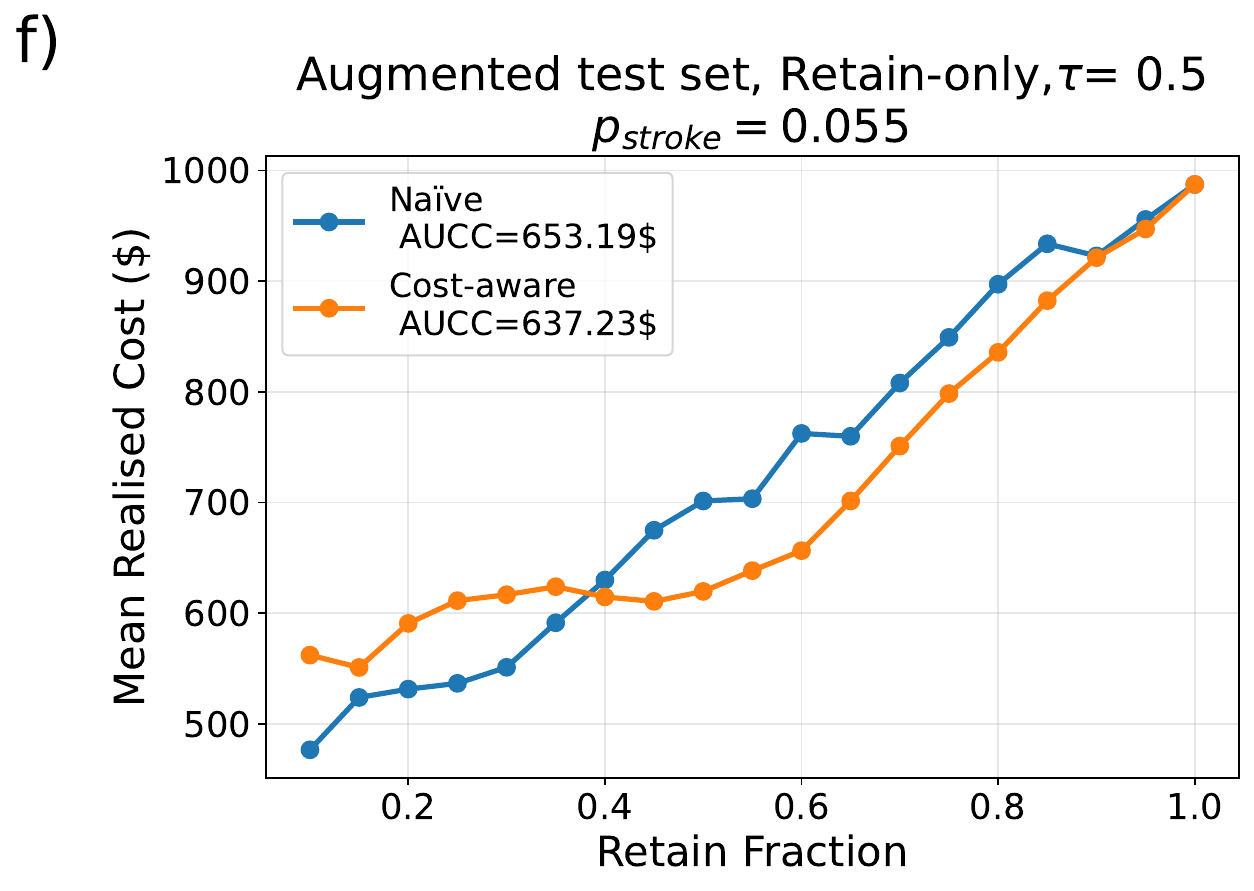}
    \caption{Cost-coverage curves and AUCC for the original and augmented test sets.}
    \label{fig:coveragecurvesnoise0105}
    
\end{figure}

\begin{table}[h]
\centering
\scriptsize
\caption{Composition of prediction outcomes for the overall test set and rejected subset at a 50\% retain fraction for $\sigma=0.1$. N denotes the na\"{i}ve rule and C denotes the cost-aware rule. Percentages are calculated relative to the total number of examples in the corresponding subset. A decision threshold of $\tau=0.5$ is used for both
the na\"{i}ve and cost-aware rules.}
\label{tab:error-composition-noise0105}
\begin{tabular}{lllrrrrr}
\toprule
$p_{\mathrm{stroke}}$ & Rule/data & Subset & Total
& TN, n (\%) & FN, n (\%) & TP, n (\%) & FP, n (\%) \\
\midrule

All & Augmented & Overall & 15377
& 7774 (50.56\%) & 3299 (21.45\%)
& 2498 (16.25\%) & 1806 (11.74\%) \\

All & Original & Overall & 15377
& 9171 (59.64\%) & 3464 (22.53\%)
& 2333 (15.17\%) & 409 (2.66\%) \\

\midrule

All & N, augmented & Rejected & 7689
& 2912 (37.87\%) & 2496 (32.46\%)
& 1222 (15.89\%) & 1059 (13.77\%) \\

All & N, original & Rejected & 7689
& 3408 (44.32\%) & 2816 (36.62\%)
& 1097 (14.27\%) & 368 (4.79\%) \\

\midrule

0.025 & C, augmented & Rejected & 7689
& 3937 (51.20\%) & 2937 (38.20\%)
& 425 (5.53\%) & 390 (5.07\%) \\

0.025 & C, original & Rejected & 7689
& 4242 (55.17\%) & 2991 (38.90\%)
& 309 (4.02\%) & 147 (1.91\%) \\

\midrule

0.037 & C, augmented & Rejected & 7689
& 4617 (60.05\%) & 3072 (39.95\%)
& 0 (0.00\%) & 0 (0.00\%) \\

0.037 & C, original & Rejected & 7689
& 4675 (60.80\%) & 3014 (39.20\%)
& 0 (0.00\%) & 0 (0.00\%) \\

\midrule

0.055 & C, augmented & Rejected & 7689
& 4643 (60.38\%) & 3046 (39.62\%)
& 0 (0.00\%) & 0 (0.00\%) \\

0.055 & C, original & Rejected & 7689
& 4675 (60.80\%) & 3014 (39.20\%)
& 0 (0.00\%) & 0 (0.00\%) \\

\bottomrule
\end{tabular}
\end{table}

\begin{table}[h] \centering \small \caption{Area under the cost--coverage curve (AUCC) for the na\"{i}ve and cost-aware filtering rules on the original and noise-augmented test sets for $\sigma=0.1$. Lower AUCC indicates lower realised cost across the evaluated retain fractions. A common decision threshold of $\tau=0.5$ is used for both the na\"{i}ve and cost-aware rules.} \label{tab:aucc-noise0105} \begin{tabular}{llrr} \toprule $p_{\mathrm{stroke}}$ & Test set & Na\"{i}ve AUCC (\$) & Cost-aware AUCC (\$) \\ \midrule 0.025 & Original & 458.6814 & 451.8226 \\ 0.025 & Augmented & 519.5305 & 527.9300 \\ \midrule 0.037 & Original & 512.4814 & 496.9960 \\ 0.037 & Augmented & 572.9958 & 579.9770 \\ \midrule 0.055 & Original & 593.1815 & 560.6090 \\ 0.055 & Augmented & 653.1937 & 637.2257 \\ \bottomrule \end{tabular} \end{table}

\newpage
\clearpage
\subsection{$\sigma=0.3$}

\subsubsection{$\tau=0.5$}
\label{sec:tau5retain}
\begin{figure}[h]
    \centering
    Noise $\sigma=0.3$, $\tau=0.5$ 
    \includegraphics[width=0.45\linewidth]{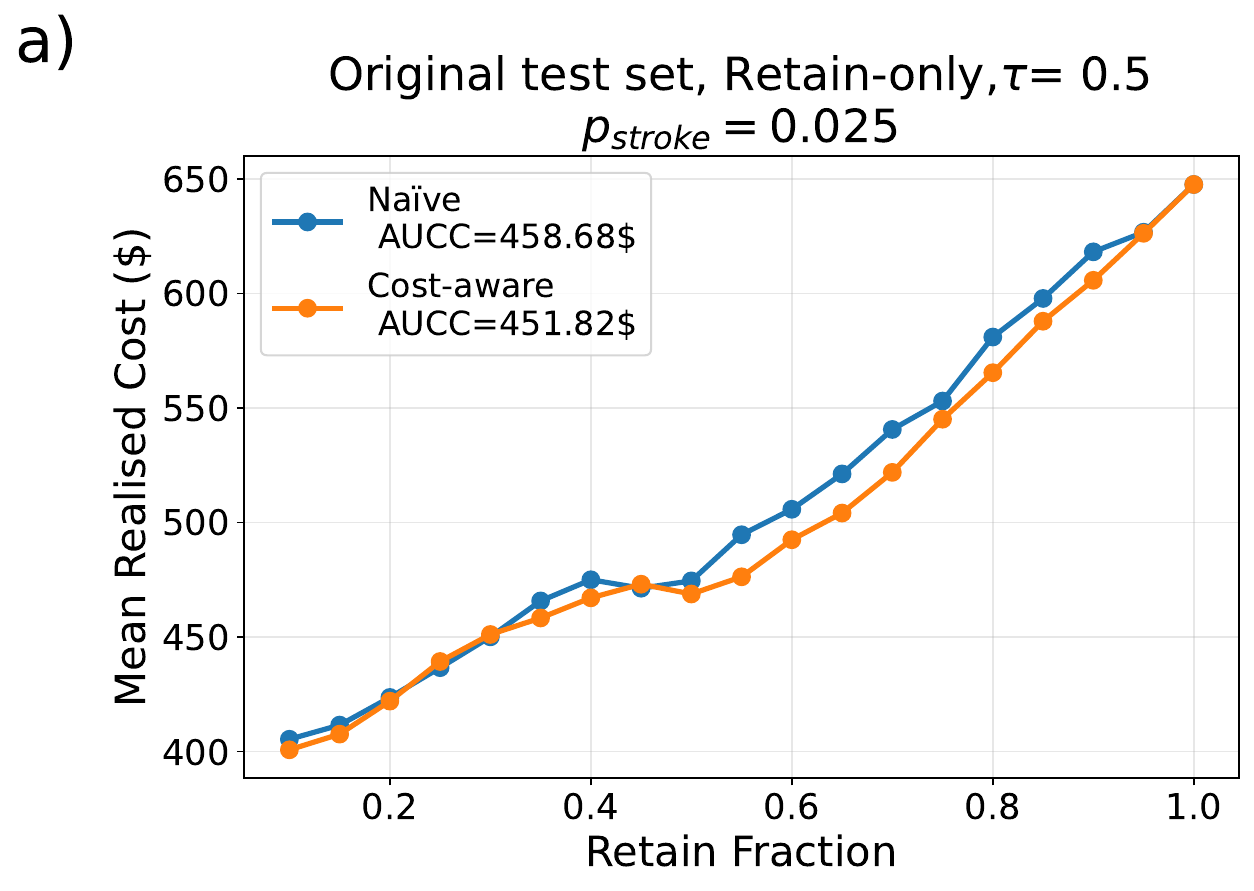}\includegraphics[width=0.45\linewidth]{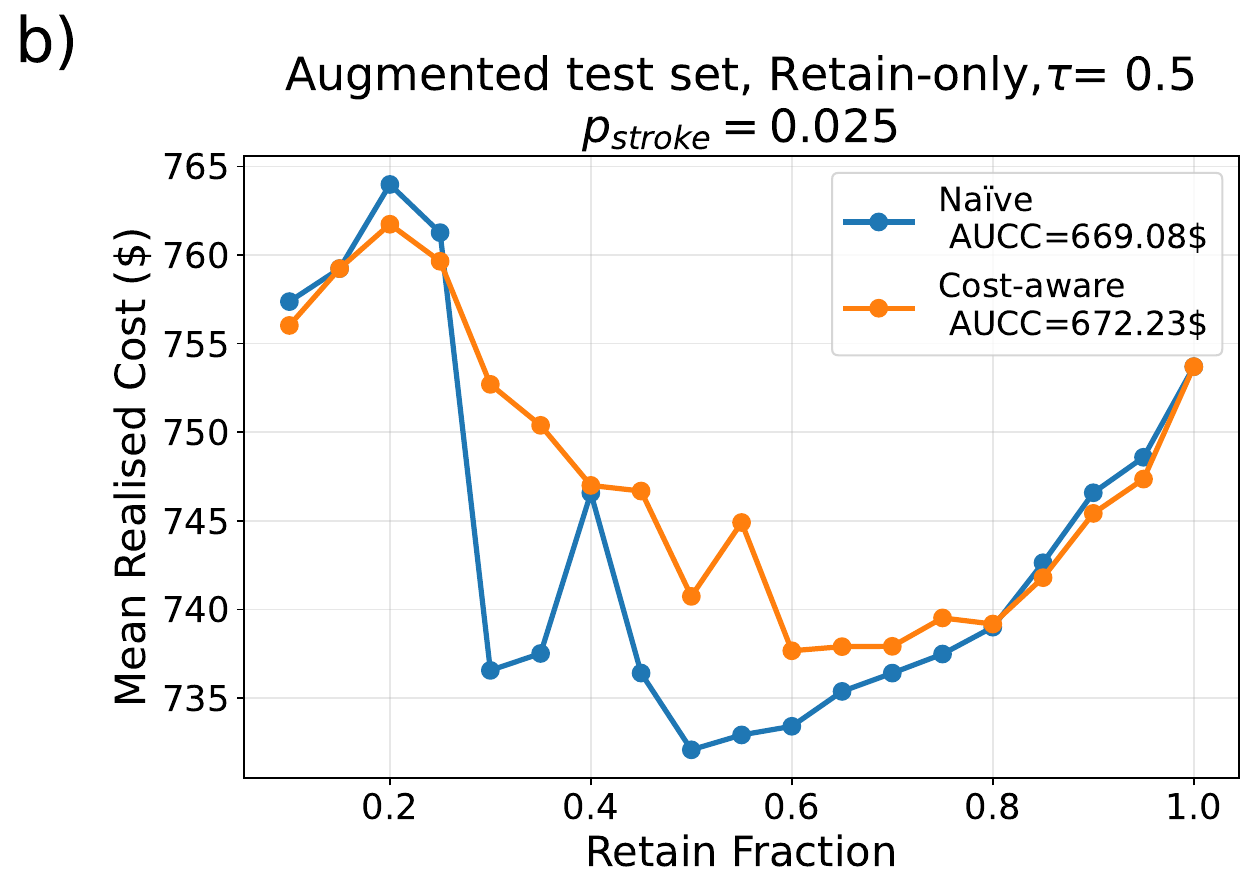}
    \includegraphics[width=0.45\linewidth]{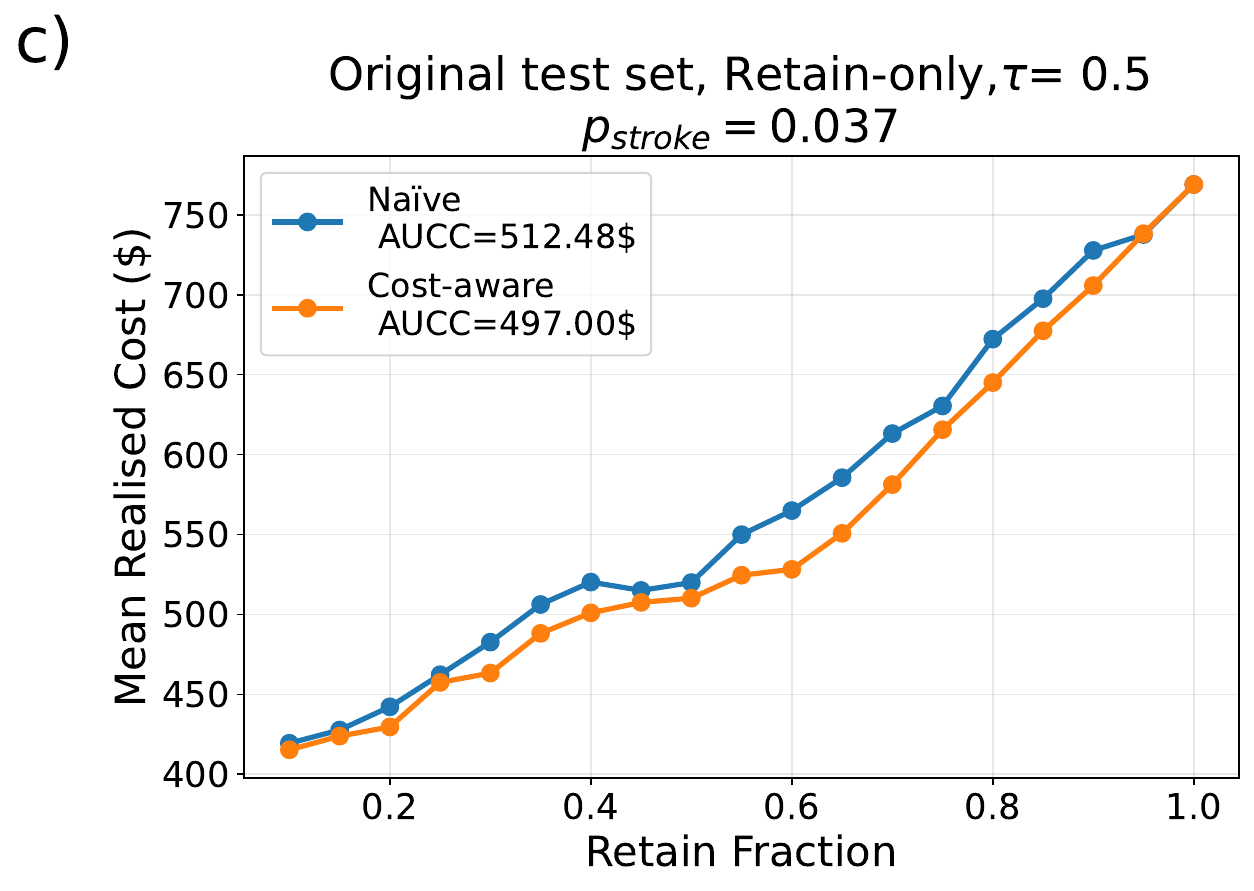}
    \includegraphics[width=0.45\linewidth]{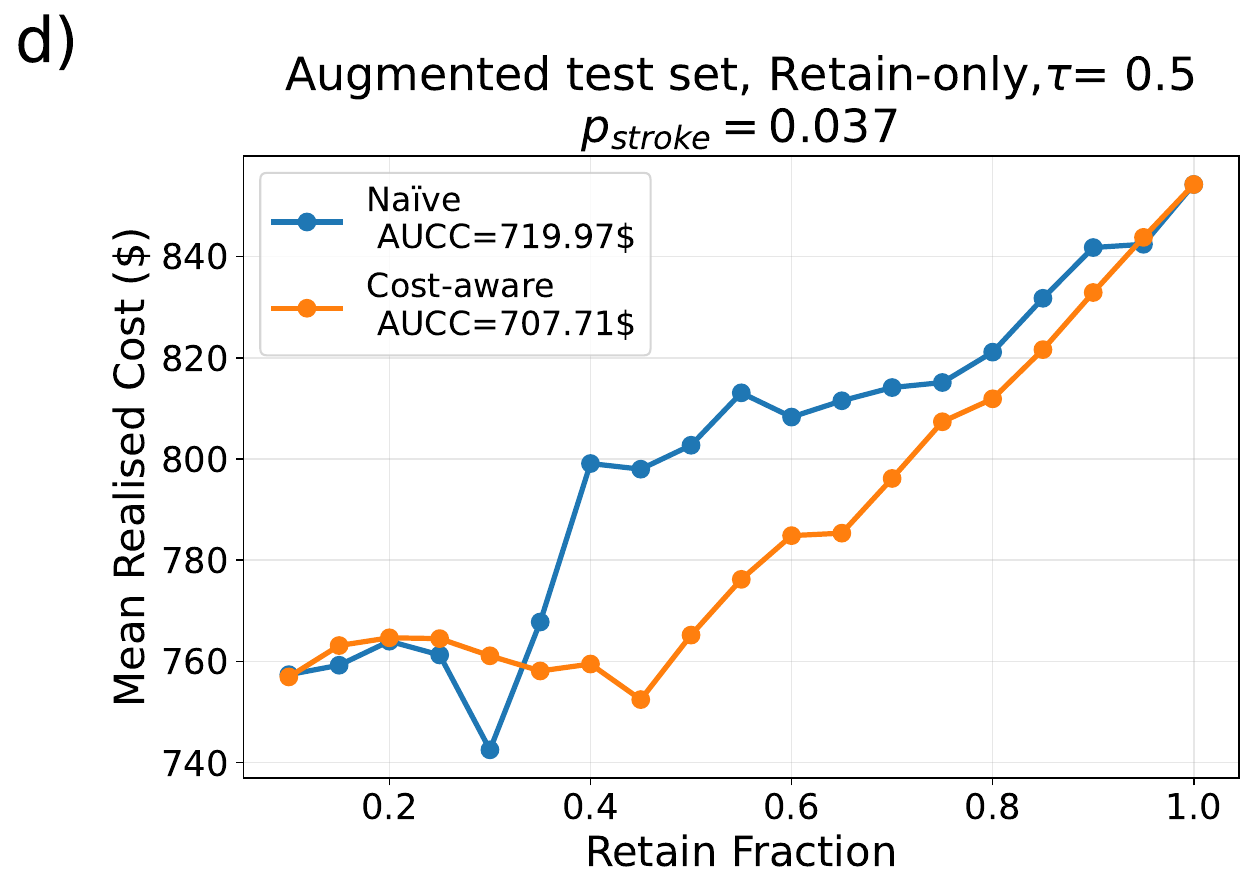}
    \includegraphics[width=0.45\linewidth]{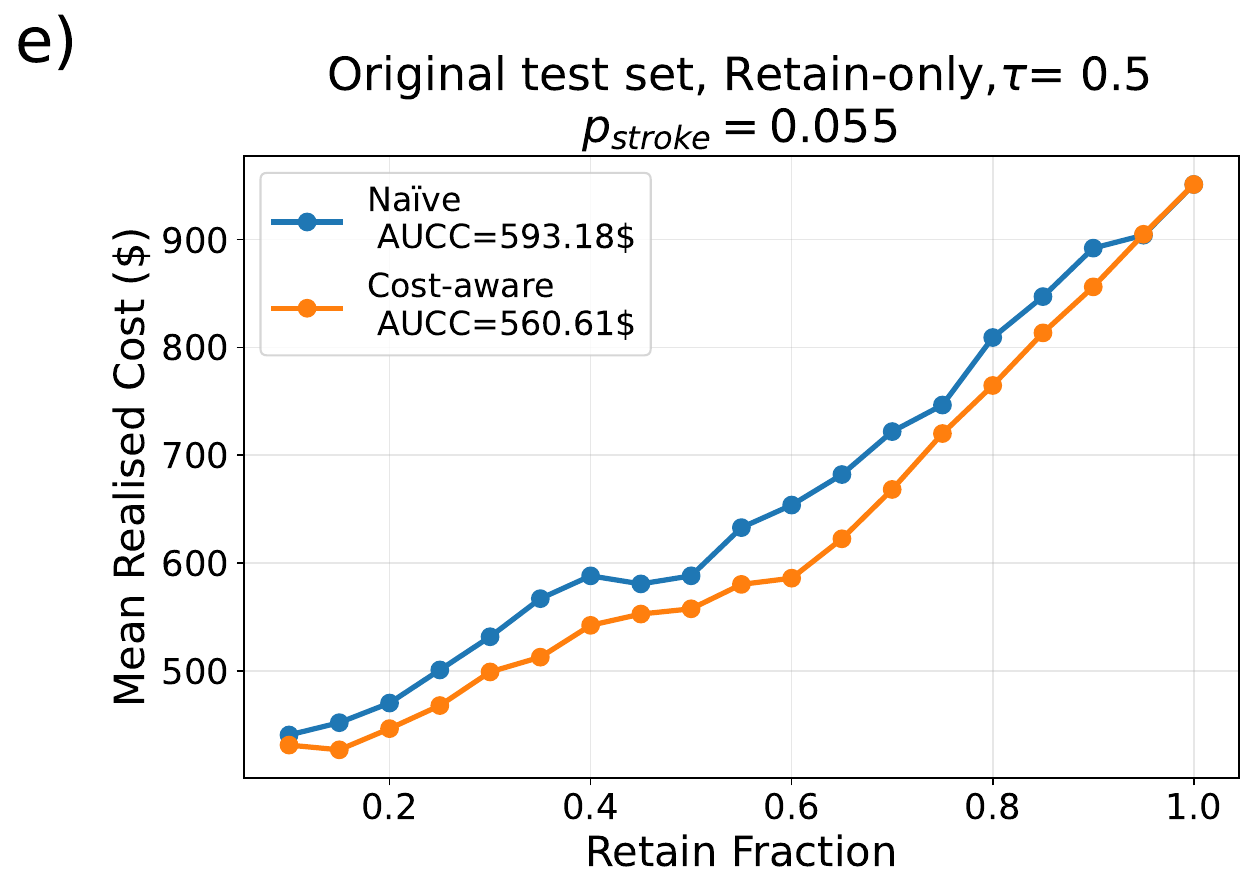}\includegraphics[width=0.45\linewidth]{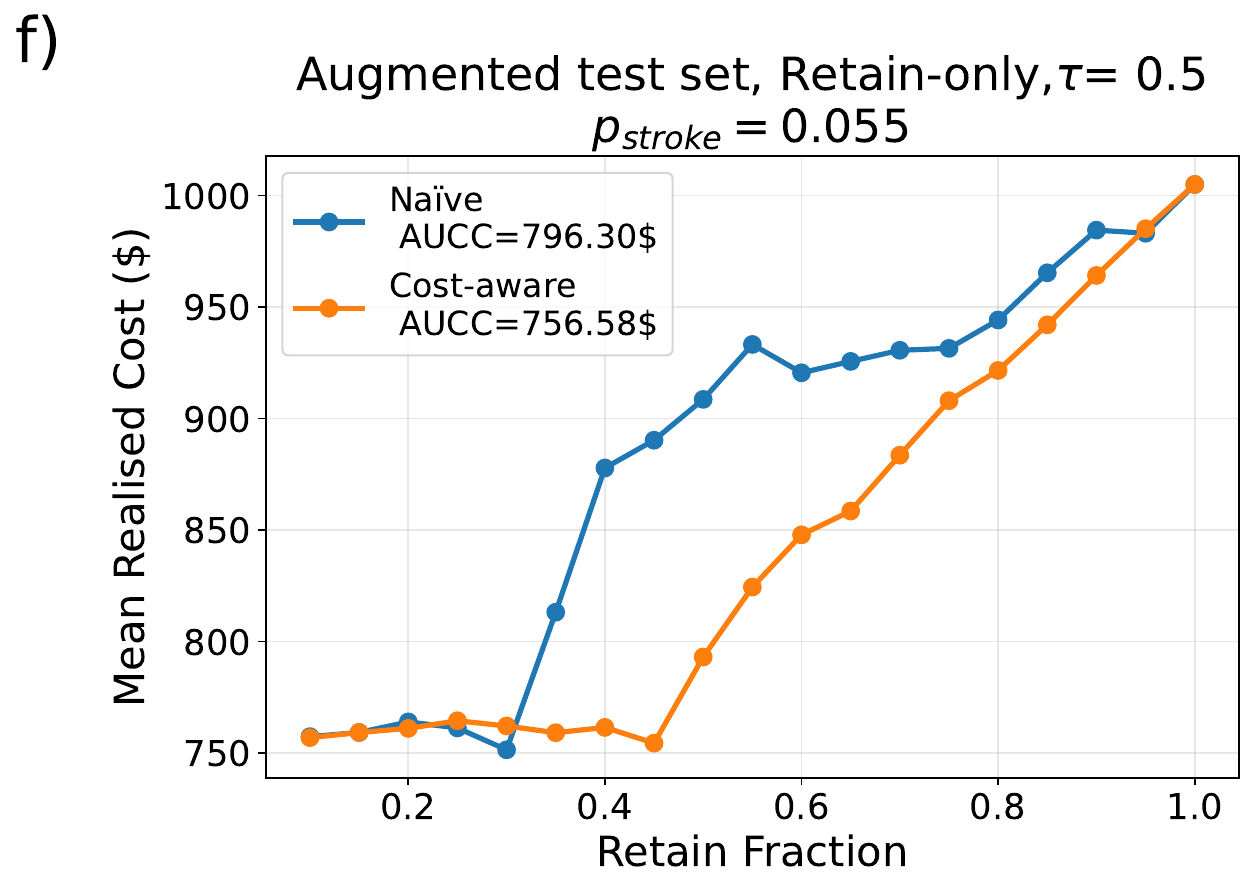}
    \caption{Cost-coverage curves and AUCC for the original and augmented test sets ($\tau=0.5$, $\sigma=0.3$).}
    \label{fig:coveragecurvesnoise0305}
\end{figure}

\begin{table}[h] \centering \scriptsize \caption{Composition of prediction outcomes for the overall test set and rejected subset at a 50\% retain fraction for $\sigma=0.3$. N denotes the na\"{i}ve rule and C denotes the cost-aware rule. Percentages are calculated relative to the total number of examples in the corresponding subset. A decision threshold of $\tau=0.5$ is used for both the na\"{i}ve and cost-aware rules.} \label{tab:error-composition-noise0305} \begin{tabular}{lllrrrrr} \toprule $p_{\mathrm{stroke}}$ & Rule/data & Subset & Total & TN, n (\%) & FN, n (\%) & TP, n (\%) & FP, n (\%) \\ \midrule All & Augmented & Overall & 15377 & 5834 (37.94\%) & 2867 (18.64\%) & 2930 (19.05\%) & 3746 (24.36\%) \\ All & Original & Overall & 15377 & 9171 (59.64\%) & 3464 (22.53\%) & 2333 (15.17\%) & 409 (2.66\%) \\ \midrule All & N, augmented & Rejected & 7689 & 3295 (42.85\%) & 1860 (24.19\%) & 1114 (14.49\%) & 1420 (18.47\%) \\ All & N, original & Rejected & 7689 & 3408 (44.32\%) & 2816 (36.62\%) & 1097 (14.27\%) & 368 (4.79\%) \\ \midrule 0.025 & C, augmented & Rejected & 7689 & 4653 (60.52\%) & 2476 (32.20\%) & 239 (3.11\%) & 321 (4.17\%) \\ 0.025 & C, original & Rejected & 7689 & 4242 (55.17\%) & 2991 (38.90\%) & 309 (4.02\%) & 147 (1.91\%) \\ \midrule 0.037 & C, augmented & Rejected & 7689 & 5066 (65.89\%) & 2623 (34.11\%) & 0 (0.00\%) & 0 (0.00\%) \\ 0.037 & C, original & Rejected & 7689 & 4675 (60.80\%) & 3014 (39.20\%) & 0 (0.00\%) & 0 (0.00\%) \\ \midrule 0.055 & C, augmented & Rejected & 7689 & 5073 (65.98\%) & 2616 (34.02\%) & 0 (0.00\%) & 0 (0.00\%) \\ 0.055 & C, original & Rejected & 7689 & 4675 (60.80\%) & 3014 (39.20\%) & 0 (0.00\%) & 0 (0.00\%) \\ \bottomrule \end{tabular} \end{table}

\begin{table}[h] \centering \small \caption{Area under the cost-coverage curve (AUCC) for the na\"{i}ve rule and the cost-aware rule on the original and augmented test sets for $\sigma=0.3$. Lower AUCC indicates lower realised cost across the evaluated retain fractions. A decision threshold of $\tau=0.5$ is used for both the na\"{i}ve and cost-aware rules.} \label{tab:aucc-noise0305} \begin{tabular}{llrr} \toprule $p_{\mathrm{stroke}}$ & Test set & Na\"{i}ve AUCC (\$) & Cost-aware AUCC (\$) \\ \midrule 0.025 & Original & 458.6814 & 451.8226 \\ 0.025 & Augmented & 669.0770 & 672.2320 \\ \midrule 0.037 & Original & 512.4814 & 496.9960 \\ 0.037 & Augmented & 719.9655 & 707.7051 \\ \midrule 0.055 & Original & 593.1815 & 560.6090 \\ 0.055 & Augmented & 796.2983 & 756.5807 \\ \bottomrule \end{tabular} \end{table}

\newpage
\clearpage
\subsection{$\sigma=0.6$}
\subsubsection{Decision optimal $\tau$}
\begin{figure}[h]
    \centering
    Noise $\sigma=0.6$, Decision optimal $\tau$ for cost-aware rule
    \includegraphics[width=0.45\linewidth]{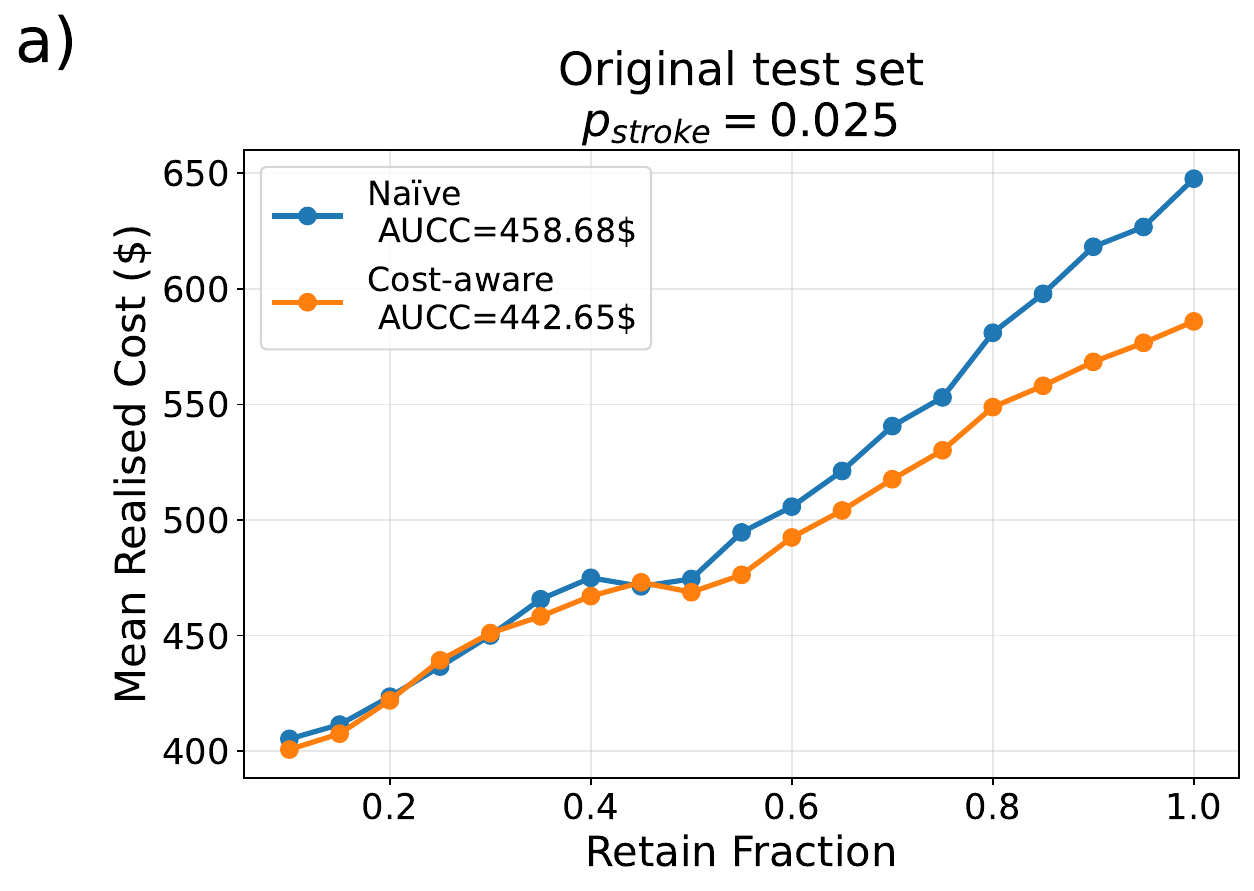}\includegraphics[width=0.45\linewidth]{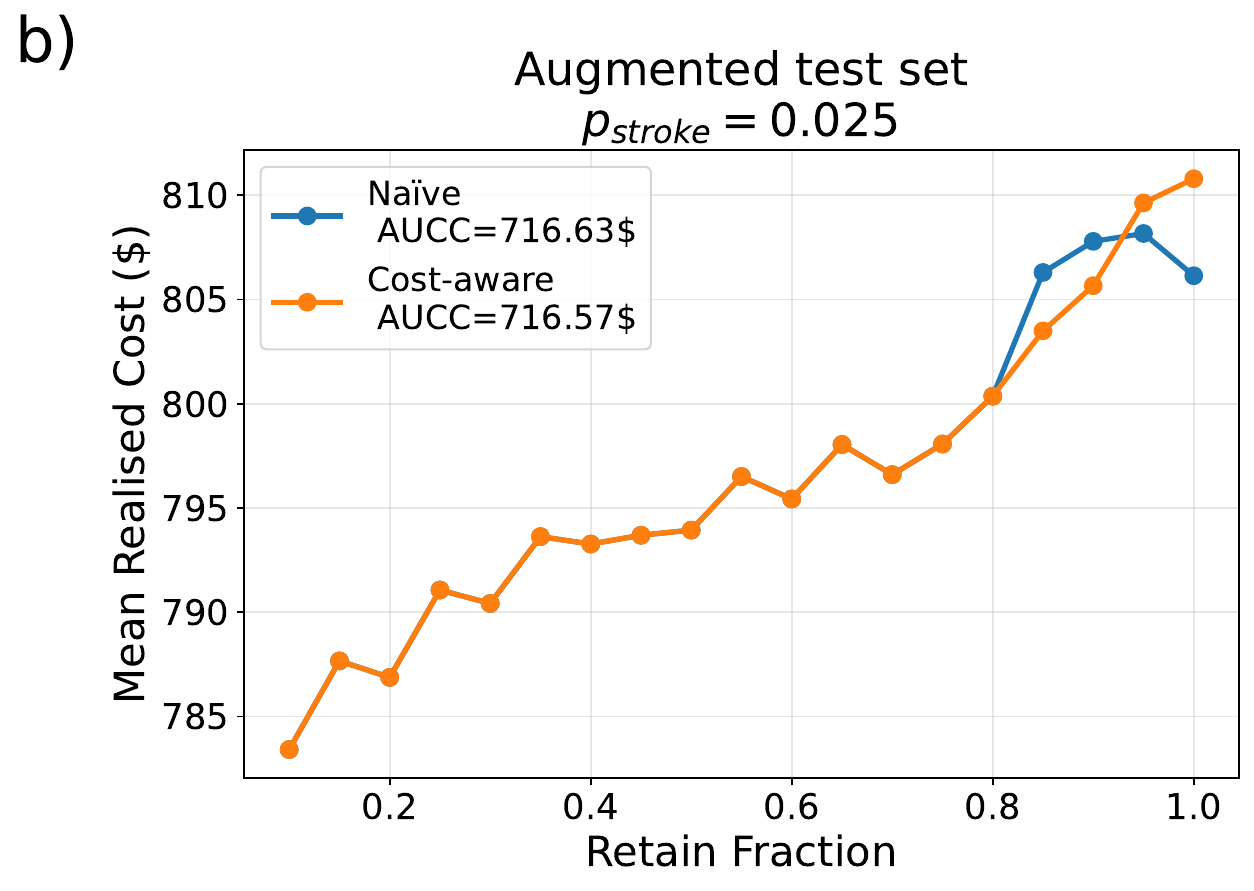}
    \includegraphics[width=0.45\linewidth]{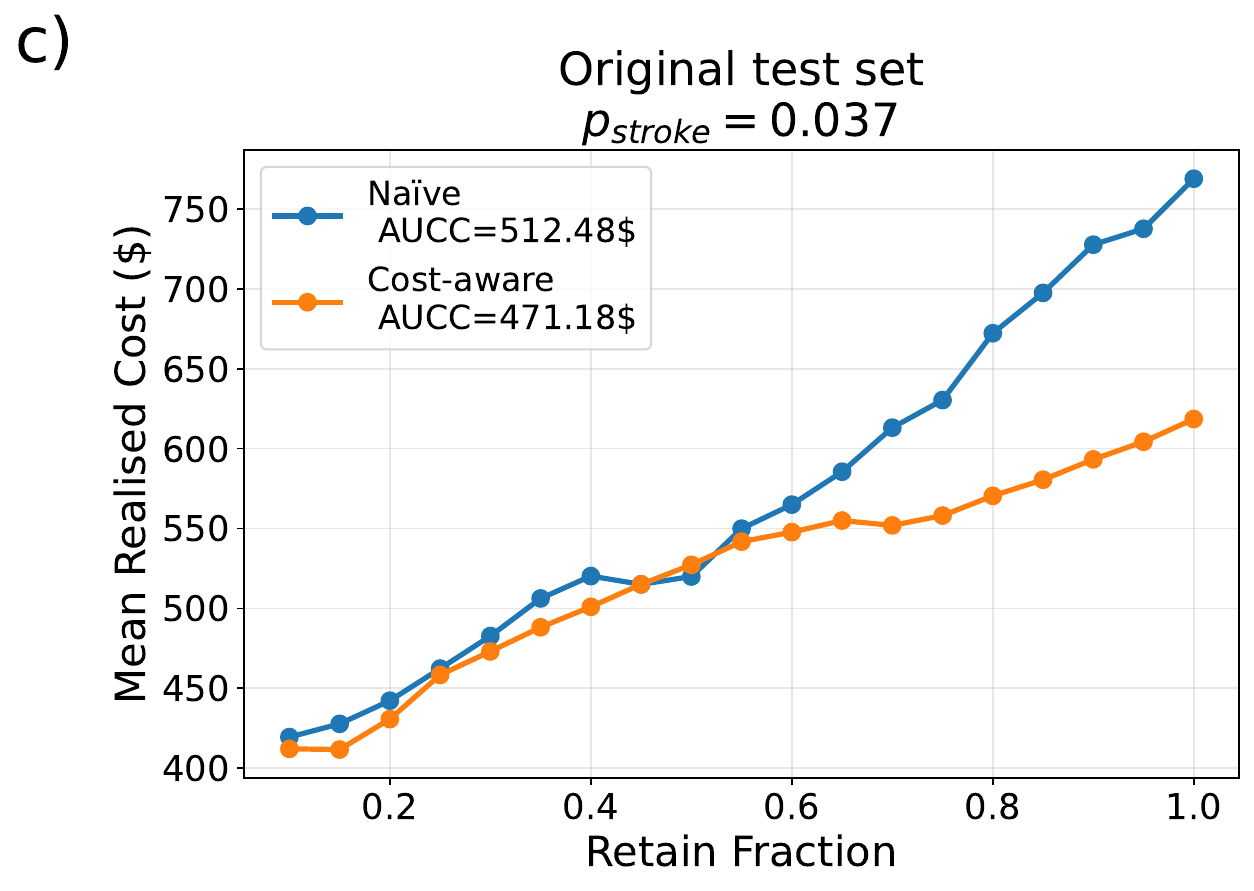}
    \includegraphics[width=0.45\linewidth]{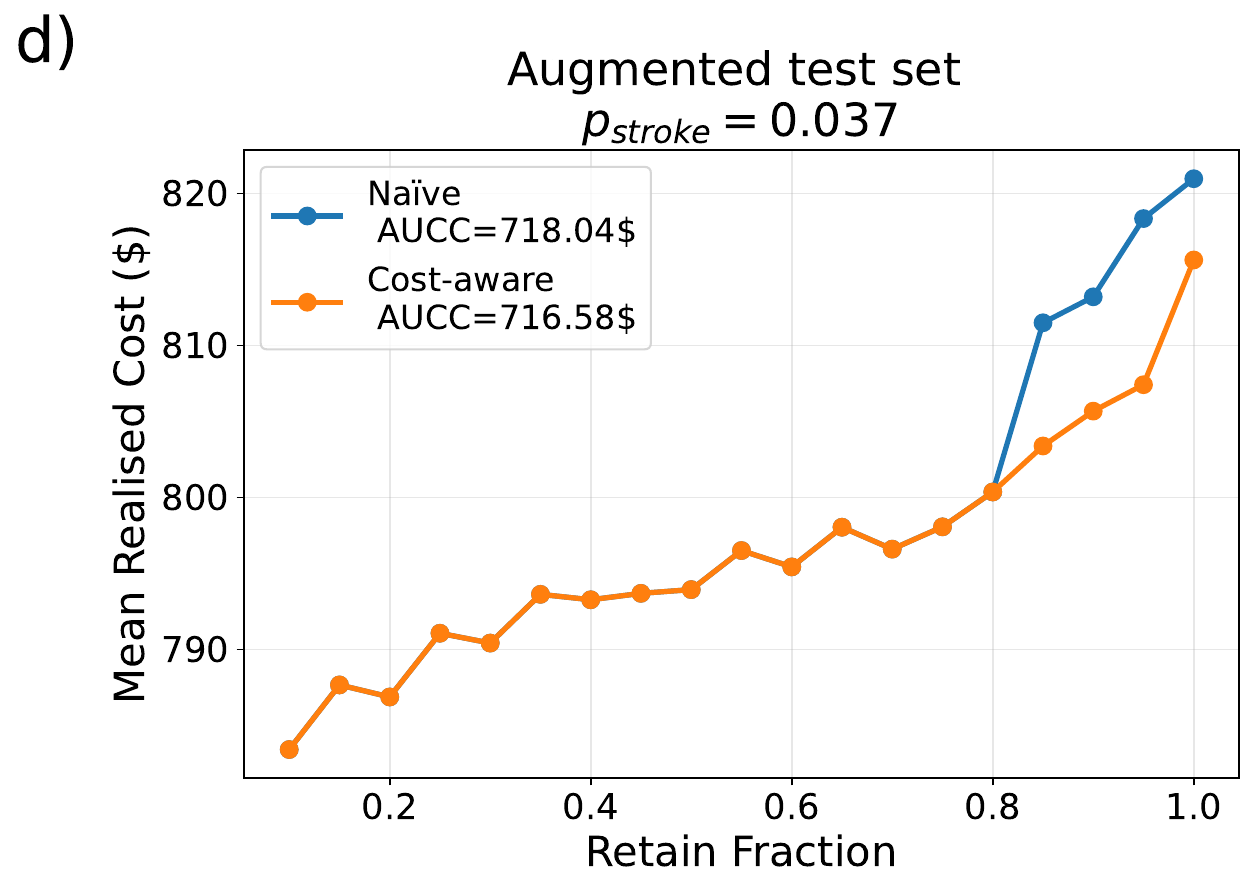}
    \includegraphics[width=0.45\linewidth]{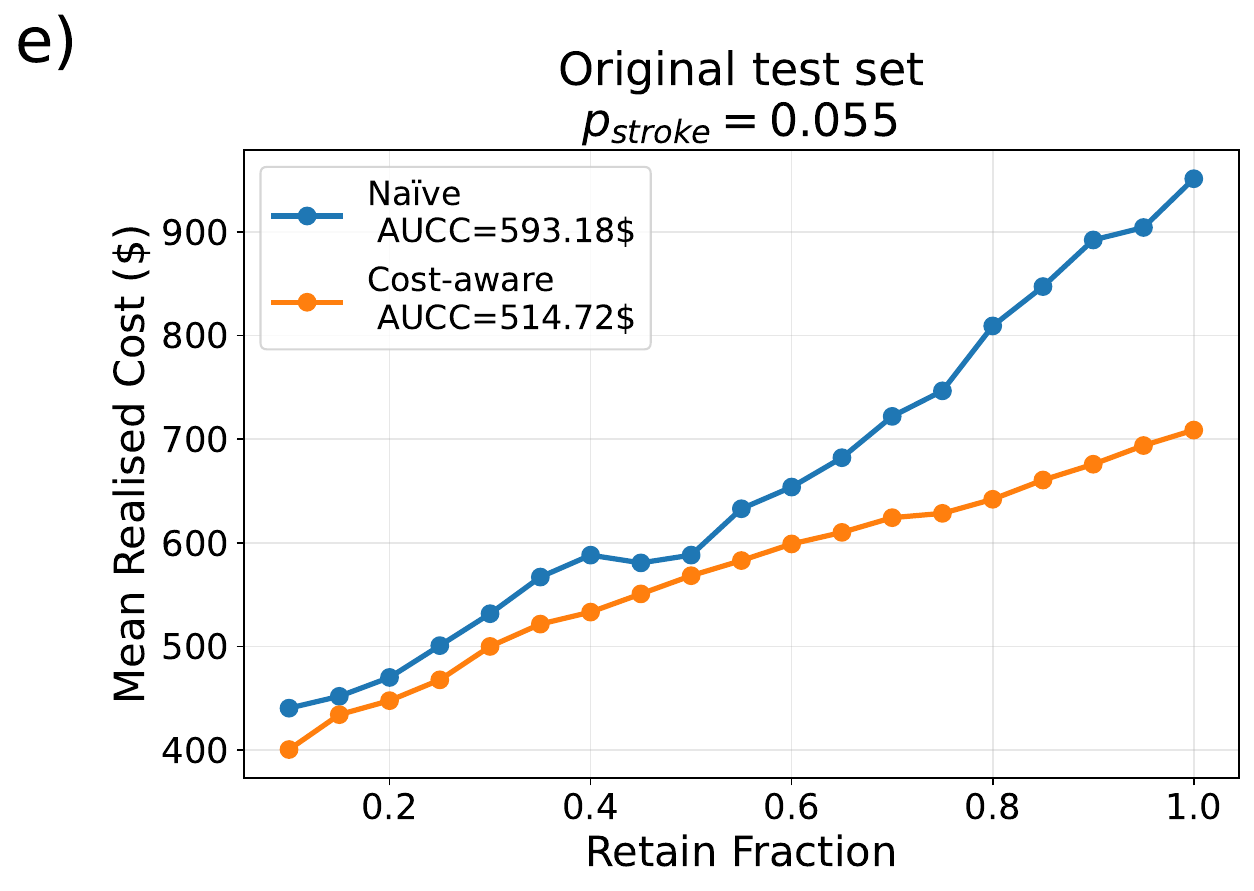}\includegraphics[width=0.45\linewidth]{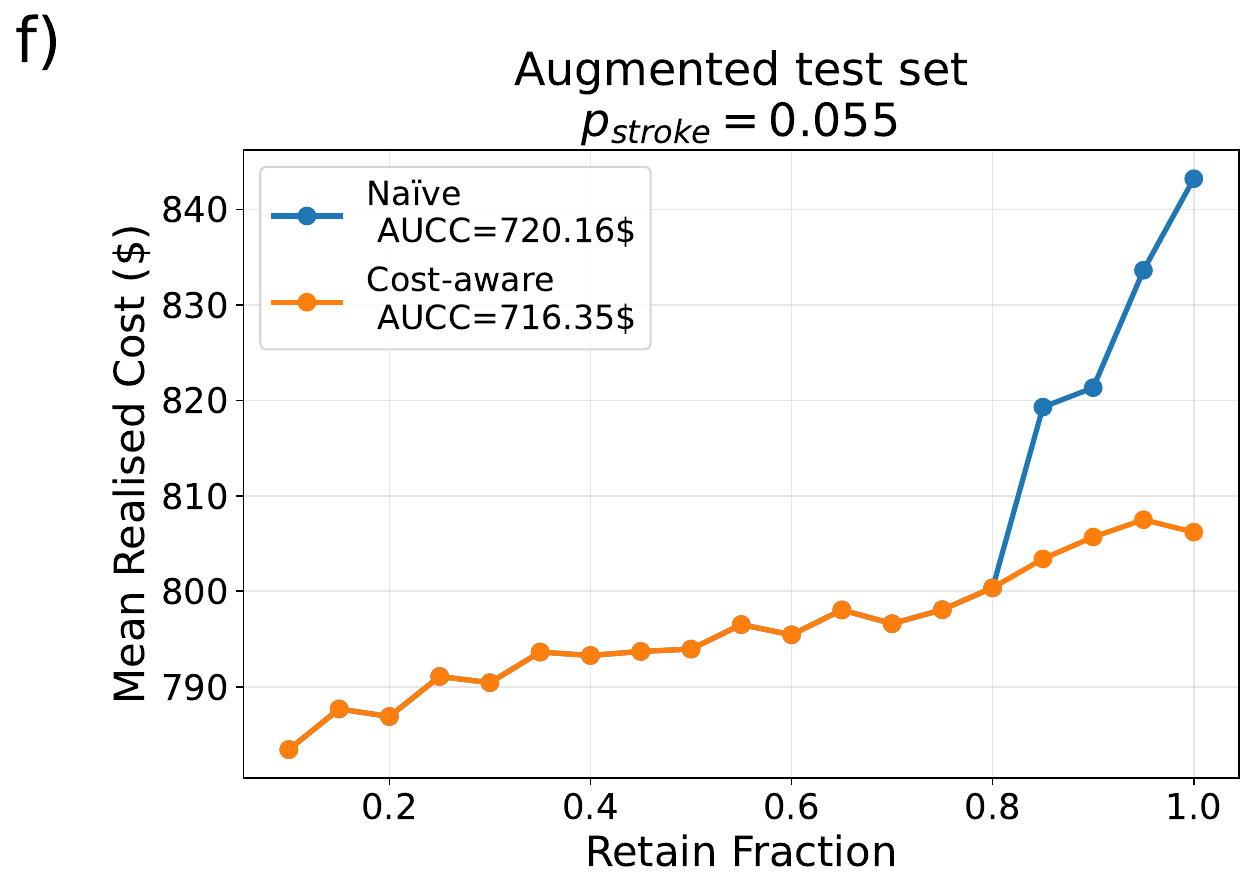}
    \caption{Cost-coverage curves and AUCC for the original and augmented test sets (decision-optimal values of $\tau$ for the cost-aware rule, and $\tau=0.5$ for the na\"{i}ve rule).}
    \label{fig:coveragecurvesnoise06}
\end{figure}
\begin{table}[h] \centering \scriptsize \caption{Composition of prediction outcomes for the overall test set and rejected subset at a 50\% retain fraction for $\sigma=0.6$. N denotes the na\"{i}ve rule and C denotes the cost-aware rule. Percentages are calculated relative to the total number of examples in the corresponding subset. We use decision-optimal values of $\tau$ for the cost-aware rule, and $\tau=0.5$ for the na\"{i}ve rule. } \label{tab:error-compositionnoise06} \begin{tabular}{lllrrrrr} \toprule $p_{\mathrm{stroke}}$ & Rule/data & Subset & Total & TN, n (\%) & FN, n (\%) & TP, n (\%) & FP, n (\%) \\ \midrule All & Augmented & Overall & 15377 & 688 (4.47\%) & 423 (2.75\%) & 5374 (34.95\%) & 8892 (57.83\%) \\ All & Original & Overall & 15377 & 9171 (59.64\%) & 3464 (22.53\%) & 2333 (15.17\%) & 409 (2.66\%) \\ \midrule All & N, augmented & Rejected & 7689 & 688 (8.95\%) & 423 (5.50\%) & 2340 (30.43\%) & 4238 (55.12\%) \\ All & N, original & Rejected & 7689 & 3408 (44.32\%) & 2816 (36.62\%) & 1097 (14.27\%) & 368 (4.79\%) \\ \midrule 0.025 & C, augmented & Rejected & 7689 & 184 (2.39\%) & 177 (2.30\%) & 2586 (33.63\%) & 4742 (61.67\%) \\ 0.025 & C, original & Rejected & 7689 & 2335 (30.37\%) & 975 (12.68\%) & 2325 (30.24\%) & 2054 (26.71\%) \\ \midrule 0.037 & C, augmented & Rejected & 7689 & 123 (1.60\%) & 139 (1.81\%) & 2624 (34.13\%) & 4803 (62.47\%) \\ 0.037 & C, original & Rejected & 7689 & 3177 (41.32\%) & 694 (9.03\%) & 1772 (23.05\%) & 2046 (26.61\%) \\ \midrule 0.055 & C, augmented & Rejected & 7689 & 2 (0.03\%) & 1 (0.01\%) & 2762 (35.92\%) & 4924 (64.04\%) \\ 0.055 & C, original & Rejected & 7689 & 731 (9.51\%) & 98 (1.27\%) & 1941 (25.24\%) & 4919 (63.97\%) \\ \bottomrule \end{tabular} \end{table}

\begin{table}[t] \centering \small \caption{Area under the cost--coverage curve (AUCC) for the na\"{i}ve and cost-aware filtering rules on the original and noise-augmented test sets for $\sigma=0.6$. Lower AUCC indicates lower realised cost across the evaluated retain fractions. The na\"{i}ve rule uses $\tau=0.5$, whereas the cost-aware rule uses the decision-optimal thresholds $\tau=\{0.381,0.293,0.218\}$ for $p_{\mathrm{stroke}}=\{0.025,0.037,0.055\}$, respectively.} \label{tab:auccnoise06} \begin{tabular}{llrr} \toprule $p_{\mathrm{stroke}}$ & Test set & Na\"{i}ve AUCC (\$) & Cost-aware AUCC (\$) \\ \midrule 0.025 & Original & 458.6814 & 442.6533 \\ 0.025 & Augmented & 716.6292 & 716.5725 \\ \midrule 0.037 & Original & 512.4814 & 471.1761 \\ 0.037 & Augmented & 718.0400 & 716.5785 \\ \midrule 0.055 & Original & 593.1815 & 514.7212 \\ 0.055 & Augmented & 720.1561 & 716.3463 \\ \bottomrule \end{tabular} \end{table}

\clearpage
\subsubsection{$\tau=0.5$}
\label{app:tau5retain06}

\begin{figure}[h!]
    \centering
    Noise $\sigma=0.6$, $\tau=0.5$
    \includegraphics[width=0.49\linewidth]{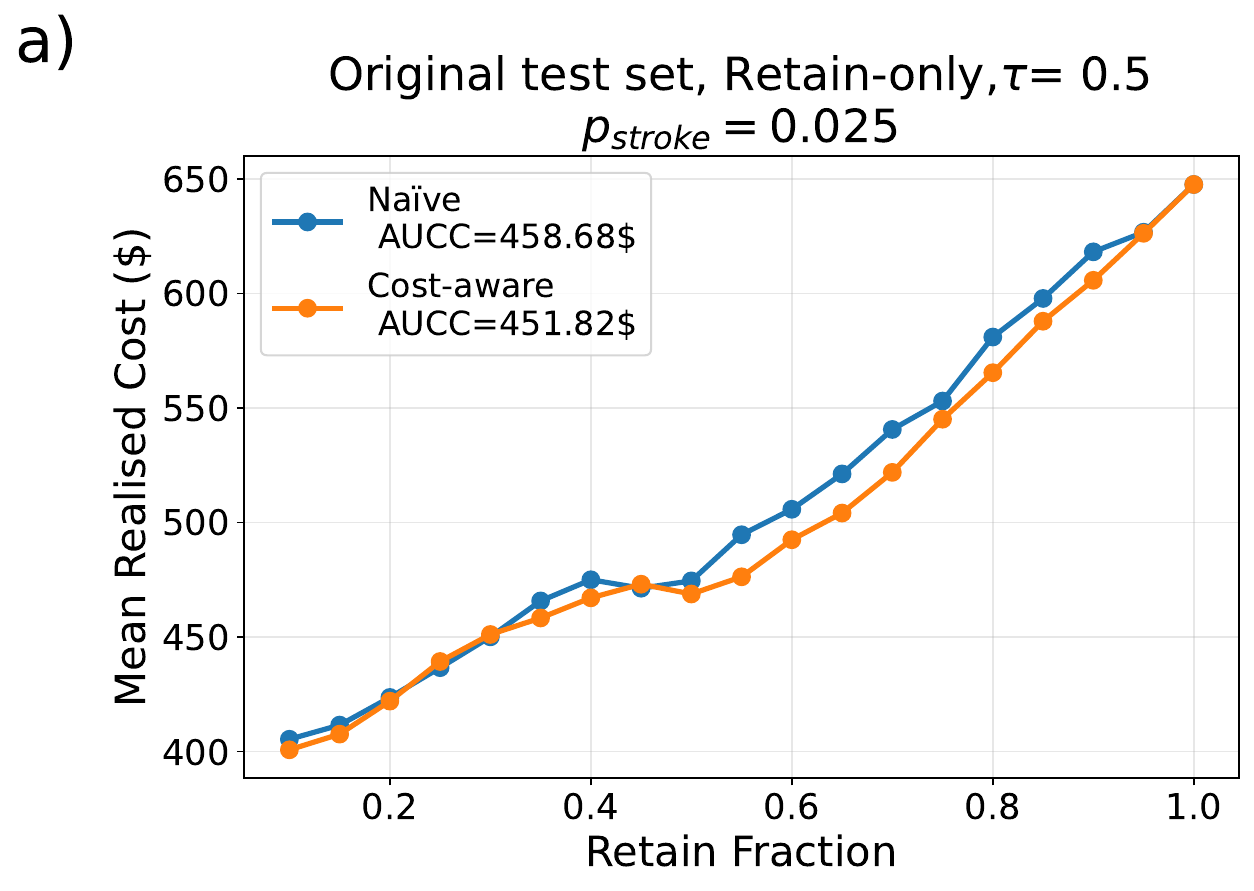}\includegraphics[width=0.49\linewidth]{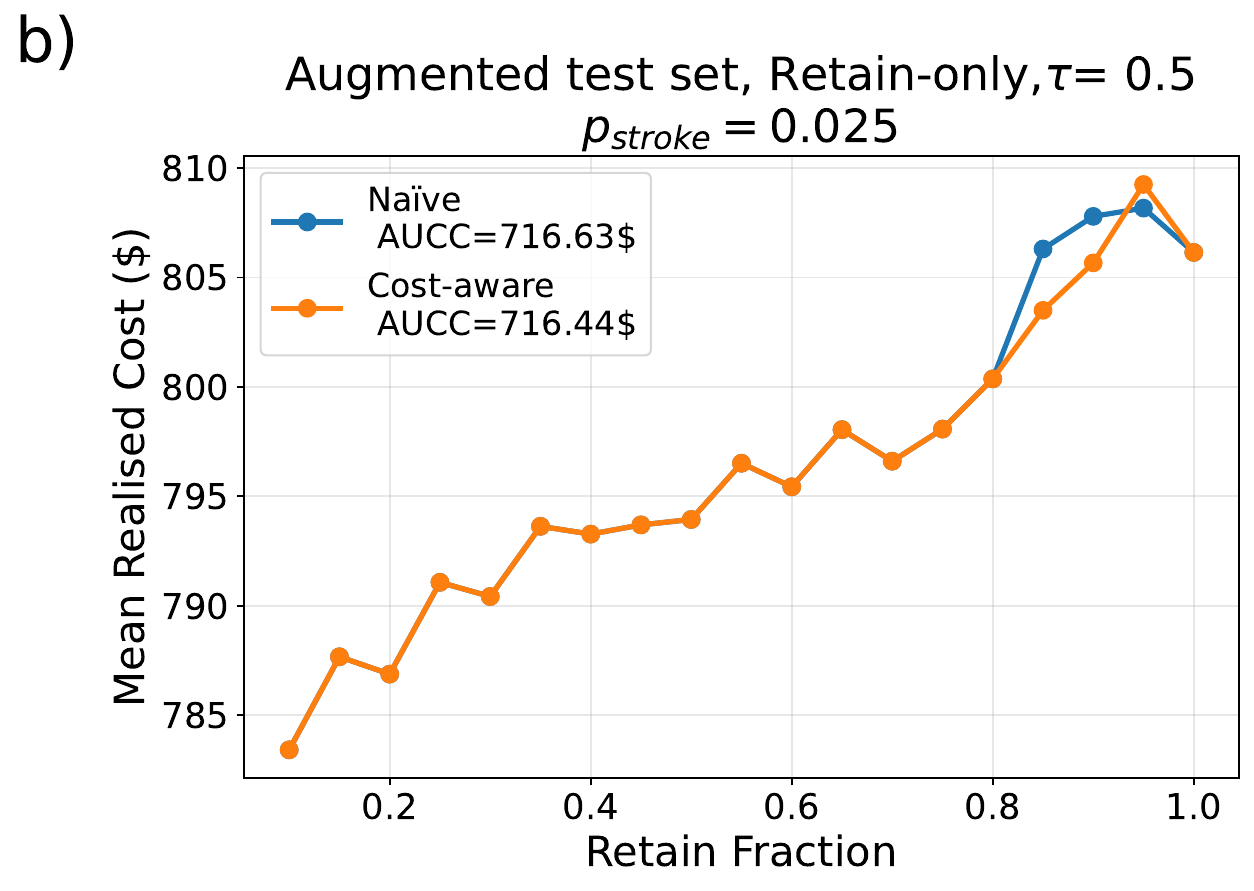}
    
    \includegraphics[width=0.49\linewidth]{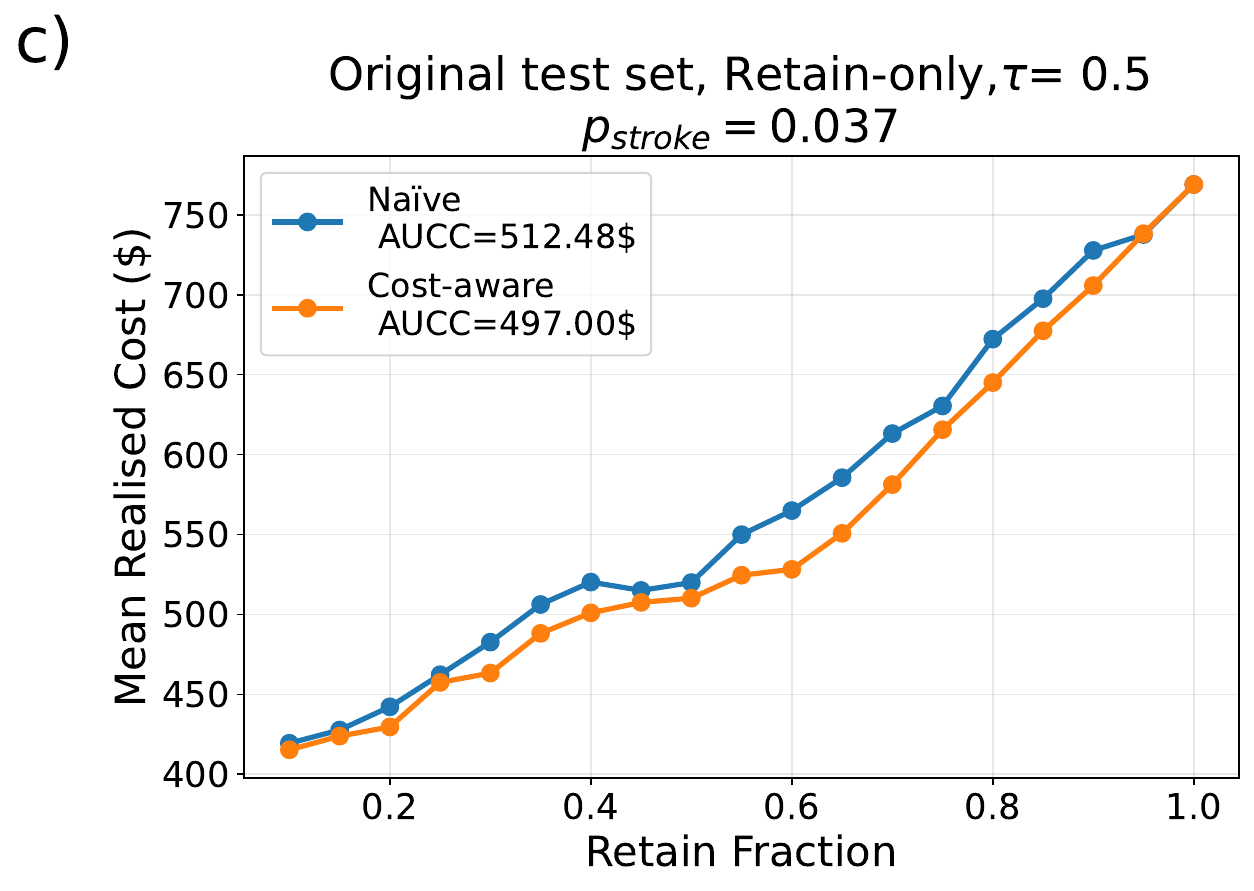}
    \includegraphics[width=0.49\linewidth]{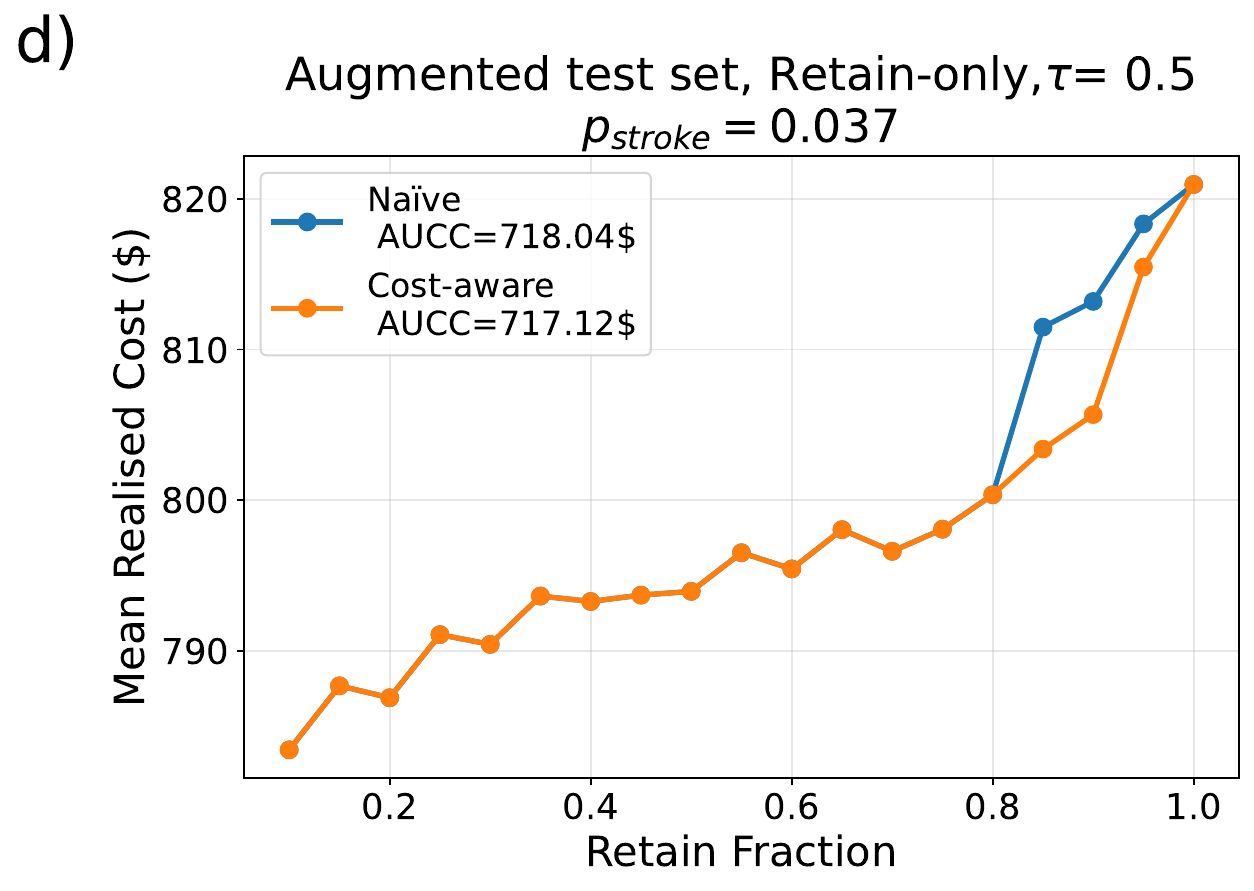}
    
    \includegraphics[width=0.49\linewidth]{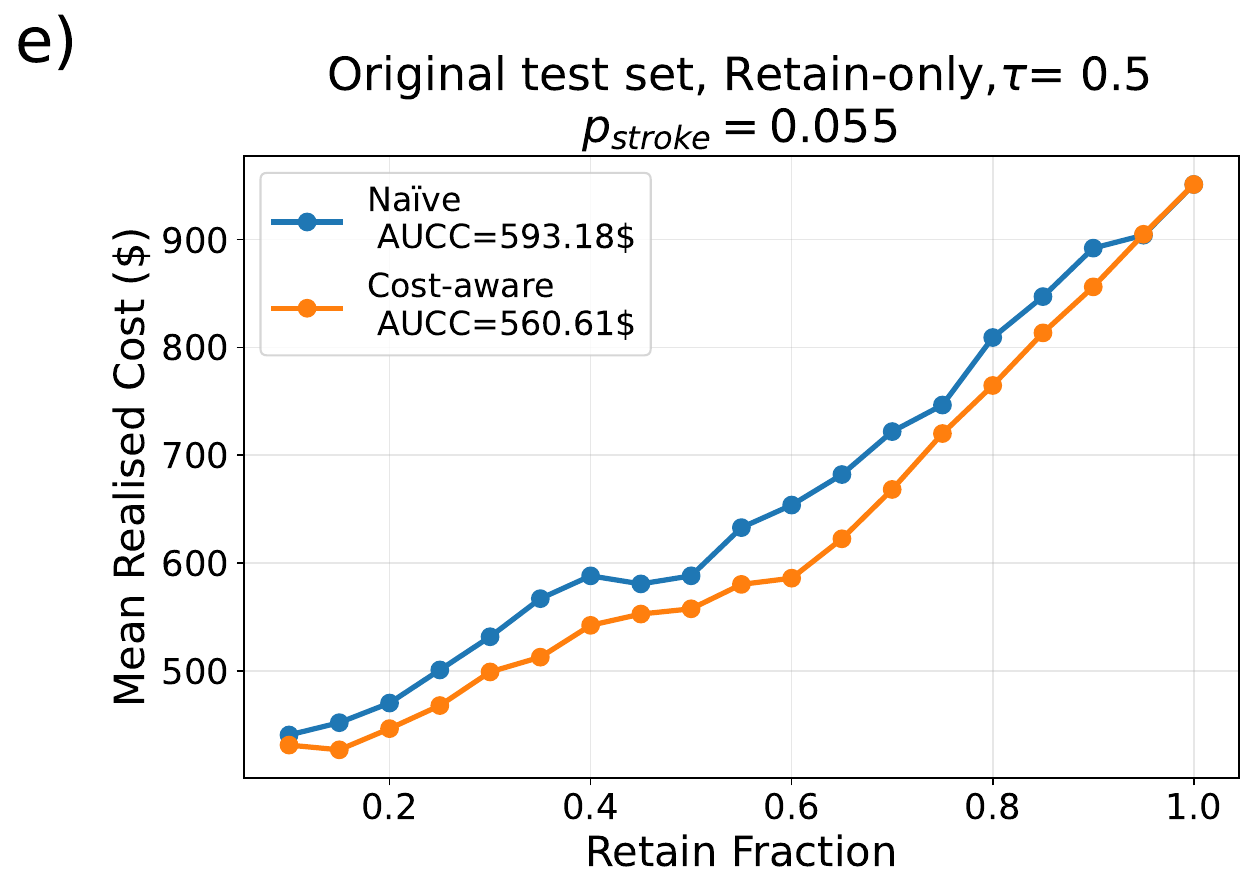}\includegraphics[width=0.49\linewidth]{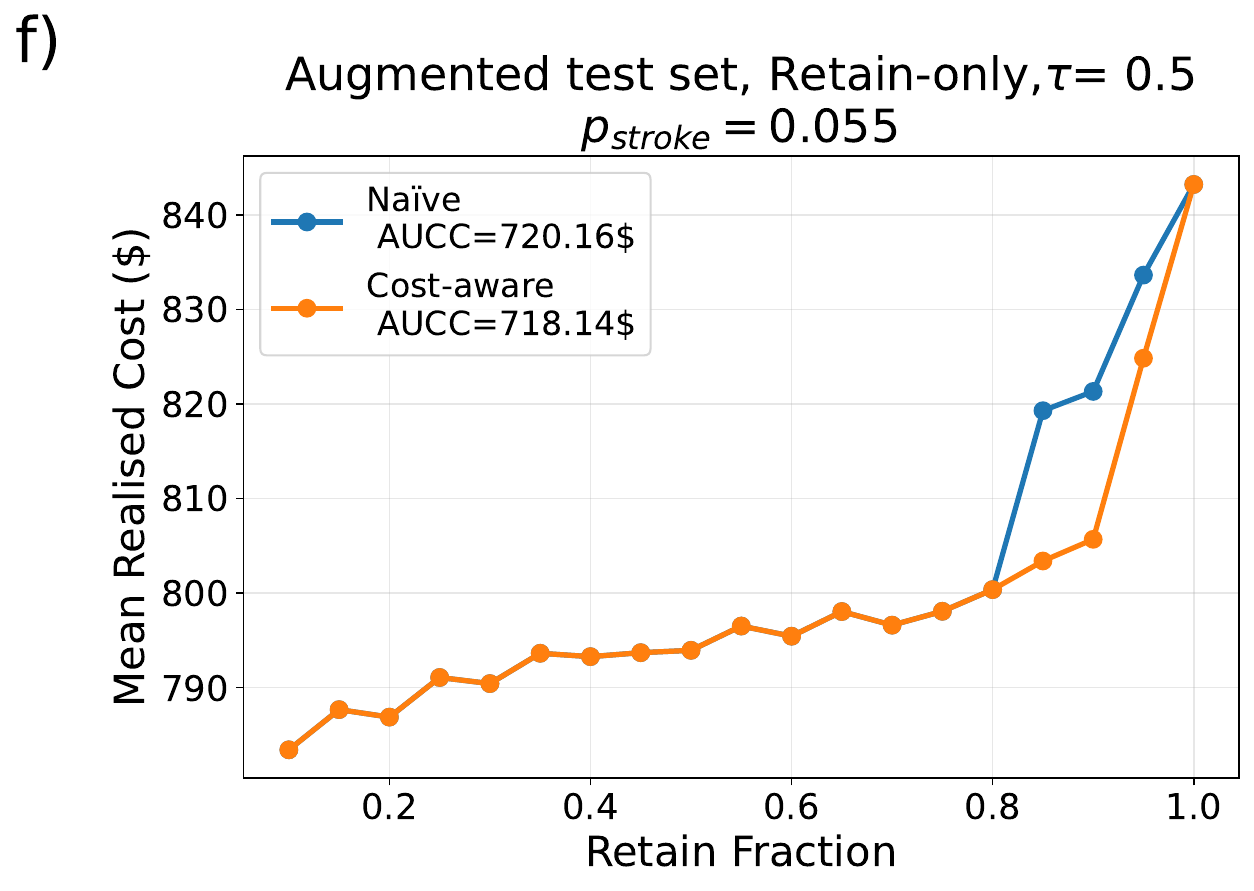}
    \caption{Cost-coverage curves with corresponding AUCC for the original and augmented test sets for $\tau=0.5$.}
    \label{fig:coveragecurvestau5}
\end{figure}

\begin{table}[h] \centering \scriptsize \caption{Composition of prediction outcomes for the overall test set and rejected subset at a 50\% retain fraction and decision threshold $\tau=0.5$ for $\sigma=0.6$. N denotes the na\"{i}ve filtering rule and C denotes the cost-aware filtering rule. The original test set is denoted non-augmented. Percentages are calculated relative to the total number of examples in the corresponding subset.} \label{tab:error-composition-threshold-05} \begin{tabular}{lllrrrrr} \toprule $p_{\mathrm{stroke}}$ & Rule/data & Subset & Total & TN, n (\%) & FN, n (\%) & TP, n (\%) & FP, n (\%) \\ \midrule All & Augmented & Overall & 15377 & 688 (4.47\%) & 423 (2.75\%) & 5374 (34.95\%) & 8892 (57.83\%) \\ All & Original & Overall & 15377 & 9171 (59.64\%) & 3464 (22.53\%) & 2333 (15.17\%) & 409 (2.66\%) \\ \midrule All & N, augmented & Rejected & 7689 & 688 (8.95\%) & 423 (5.50\%) & 2340 (30.43\%) & 4238 (55.12\%) \\ All & N, original & Rejected & 7689 & 3408 (44.32\%) & 2816 (36.62\%) & 1097 (14.27\%) & 368 (4.79\%) \\ \midrule 0.025 & C, augmented & Rejected & 7689 & 688 (8.95\%) & 423 (5.50\%) & 2340 (30.43\%) & 4238 (55.12\%) \\ 0.025 & C, original & Rejected & 7689 & 4242 (55.17\%) & 2991 (38.90\%) & 309 (4.02\%) & 147 (1.91\%) \\ \midrule 0.037 & C, augmented & Rejected & 7689 & 688 (8.95\%) & 423 (5.50\%) & 2340 (30.43\%) & 4238 (55.12\%) \\ 0.037 & C, original & Rejected & 7689 & 4675 (60.80\%) & 3014 (39.20\%) & 0 (0.00\%) & 0 (0.00\%) \\ \midrule 0.055 & C, augmented & Rejected & 7689 & 688 (8.95\%) & 423 (5.50\%) & 2340 (30.43\%) & 4238 (55.12\%) \\ 0.055 & C, original & Rejected & 7689 & 4675 (60.80\%) & 3014 (39.20\%) & 0 (0.00\%) & 0 (0.00\%) \\ \bottomrule \end{tabular} \end{table}

\begin{table}[h] \centering \small \caption{Area under the cost-coverage curve (AUCC) for the na\"{i}ve and cost-aware filtering rules on the original and noise-augmented test sets for $\sigma=0.6$. Lower AUCC indicates lower realised cost across the evaluated retain fractions. A common decision threshold of $\tau=0.5$ is used for both the na\"{i}ve and cost-aware rules.} \label{tab:aucc-threshold-05} \begin{tabular}{llrr} \toprule $p_{\mathrm{stroke}}$ & Test set & Na\"{i}ve AUCC (\$) & Cost-aware AUCC (\$) \\ \midrule 0.025 & Original & 458.6814 & 451.8226 \\ 0.025 & Augmented & 716.6292 & 716.4369 \\ \midrule 0.037 & Original & 512.4814 & 496.9960 \\ 0.037 & Augmented & 718.0400 & 717.1152 \\ \midrule 0.055 & Original & 593.1815 & 560.6090 \\ 0.055 & Augmented & 720.1561 & 718.1392 \\ \bottomrule \end{tabular} \end{table}

\clearpage
\section{Additional information about AF classification} 
\label{sec:afclass}
The DeepBeat dataset contains 25 s PPG segments sampled at 32 Hz. We use a custom subject-level split with no participant overlap and approximately consistent AF prevalence across partitions as reported in \cite{bench2026systematic}. The training set contains 40,603 AF and 65,646 non-AF segments from 88 participants, the validation set contains 5,800 AF and 9,456 non-AF segments from 26 participants, and the calibration contains 5,808 AF and 9,273 non-AF segments from 34 participants, while the test set contains 5,797 AF and 9,580 non-AF segments from 24 participants.

The original DeepBeat data were obtained through Synapse under the associated non-commercial data-use agreement. The use of the data in this study was restricted to authorised non-commercial research. Software adpated from the DeepBeat GitHub repository are subject to the GPL-3.0 licence.

The 1D AlexNet classifier was trained using stochastic gradient descent with an initial learning rate of $0.001$ and a batch size of 64. Weighted random sampling was implemented, while $T=100$ logit samples were used to calculate the training loss. Training used early stopping with a patience of 25 epochs, and the checkpoint with the lowest validation loss was retained for evaluation. The software repository contains any remaining details. Some of the code used here was modified from that reported in \cite{bench2025uncertainty}, under the terms of the Apache License 2.0.

\section{Bootstrapping analysis} \label{sec:bootstrap} 

To characterise the variability of the AUCC estimates under repeated sampling of test observations the AUCC was quantified using a paired
non-parametric bootstrap, where the original and augmented test sets were analysed separately. This shows how sensitive the estimated AUCCs, and especially the difference between the two filtering rules, are to the particular examples included in the calculation. For each experiment for $\sigma=0.3$, 1,000 bootstrap samples were generated by sampling 15,377 observation indices with replacement from the test set. Resampling was performed at the observation level. The resulting intervals characterise stability across resampled PPG segments and should not be interpreted as confidence intervals for variation across independently sampled participants.

For each bootstrap replicate, the difference in AUCC is defined as, $\Delta\mathrm{AUCC}=\mathrm{AUCC}_{\mathrm{cost\text{-}aware}}-\mathrm{AUCC}_{\text{naïve}}$.
95\% confidence intervals for the na\"ive AUCC, cost-aware AUCC, and paired AUCC difference were calculated from their respective bootstrap distributions. We interpret a 95\% bootstrap interval excluding zero as evidence that the AUCCs differed in that test-set and cost-model condition. 

For the decision optimal thresholds in Table \ref{tab:aucc_bootstrap_decision_thresholds}, the paired 95\% bootstrap confidence intervals excluded zero in all six comparisons, providing evidence against equal AUCC between the strategies. Five comparisons favoured the cost-aware strategy, whereas the noise-augmented $p_{\mathrm{stroke}}=0.025$ comparison favoured the na\"ive strategy. 

For $\tau=0.5$ in Table \ref{tab:aucc_bootstrap_common_threshold}, the paired-difference confidence intervals excluded zero in all six comparisons, so there was evidence of a difference between the filtering rules in every setting. Cost-aware filtering reduced AUCC in five comparisons but increased AUCC in the noise-augmented $p_{\mathrm{stroke}}=0.025$ comparison.

\begin{table}[h] \centering \small \setlength{\tabcolsep}{4pt} \caption{Paired empirical-bootstrap comparison of area under the cost--coverage curve (AUCC) in the decision-threshold analysis for $\sigma=0.3$. The values here correspond to the case where the na\"{i}ve strategy used the conventional classification threshold of $\tau=0.5$, whereas the cost-aware strategy used the decision-optimal thresholds $\tau=0.381$, $0.293$, and $0.218$ for $p_{\mathrm{stroke}}=0.025$, $0.037$, and $0.055$, respectively. Values in parentheses are percentile 95\% confidence intervals from 1,000 paired bootstrap replicates. The difference is defined as $\Delta\mathrm{AUCC} =\mathrm{AUCC}_{\mathrm{cost\text{-}aware}} -\mathrm{AUCC}_{\text{naïve}}$.} \label{tab:aucc_bootstrap_decision_thresholds} {\scriptsize \begin{tabular}{llrrrr} \toprule Test set & $p_{\mathrm{stroke}}$ & Na\"{i}ve AUCC & Cost-aware AUCC & $\Delta$AUCC \\ \midrule Original & 0.025 & 458.68 (452.28, 462.58) & 442.65 (437.94, 447.61) & $-16.03$ ($-16.44$, $-12.80$)\\ Original & 0.037 & 512.48 (502.95, 518.06) & 471.18 (464.98, 476.82) & $-41.31$ ($-44.54$, $-35.46$) \\ Original & 0.055 & 593.18 (579.21, 601.46) & 514.72 (507.49, 522.68) & $-78.46$ ($-84.96$, $-66.80$) \\ \midrule Noise-augmented & 0.025 & 669.08 (661.43, 674.91) & 672.28 (666.45, 678.76) & $3.20$ ($2.15$, $6.31$) \\ Noise-augmented & 0.037 & 719.97 (710.47, 727.17) & 692.72 (688.88, 701.75) & $-27.25$ ($-28.71$, $-19.17$) \\ Noise-augmented & 0.055 & 796.30 (783.85, 805.90) & 693.54 (687.66, 700.10) & $-102.76$ ($-111.84$, $-90.80$) \\ \bottomrule \end{tabular} } \end{table}

\begin{table}[h] \centering \small \setlength{\tabcolsep}{4pt} \caption{Paired empirical-bootstrap comparison of area under the cost--coverage curve (AUCC) using a common classification threshold of $\tau=0.5$. Values in parentheses are percentile 95\% confidence intervals from 1,000 paired bootstrap replicates. The difference is defined as $\Delta\mathrm{AUCC} =\mathrm{AUCC}_{\mathrm{cost\text{-}aware}} -\mathrm{AUCC}_{\text{naïve}}$. The bootstrap distribution for $p_\text{stroke}=0.025$ is biased or asymmetric enough that its central 95\% interval narrowly excludes the observed estimate. This can occur because the percentile interval is defined by quantiles of the bootstrap distribution and is not constrained to contain the observed point estimate. The particular composition of the observed test set produced a slightly larger cost-aware advantage than is typical across bootstrap resample, but direction is statistically supported.} \label{tab:aucc_bootstrap_common_threshold} {\scriptsize\begin{tabular}{llrrr} \toprule Test set & $p_{\mathrm{stroke}}$ & Na\"{i}ve AUCC & Cost-aware AUCC & $\Delta$AUCC  \\ \midrule Original & 0.025 & 458.68 (452.28, 462.58) & 451.82 (446.75, 456.74) & $-6.86$ ($-6.85$, $-4.33$)  \\ Original & 0.037 & 512.48 (502.95, 518.06) & 497.00 (489.38, 502.82) & $-15.49$ ($-16.79$, $-12.22$) \\ Original & 0.055 & 593.18 (579.21, 601.46) & 560.61 (549.82, 568.71) & $-32.57$ ($-34.38$, $-27.66$)  \\ \midrule Noise-augmented & 0.025 & 669.08 (661.43, 674.91) & 672.23 (666.24, 679.01) & $3.16$ ($2.30$, $6.02$)  \\ Noise-augmented & 0.037 & 719.97 (710.47, 727.17) & 707.71 (700.55, 714.61) & $-12.26$ ($-14.60$, $-8.44$)  \\ Noise-augmented & 0.055 & 796.30 (783.85, 805.90) & 756.58 (747.88, 765.18) & $-39.72$ ($-42.78$, $-34.18$)  \\ \bottomrule \end{tabular} }\end{table}

\clearpage

\end{document}